\documentclass{article} %
\usepackage{iclr2019_conference,times}

\usepackage{amsmath,amsfonts,bm}

\def\eqref#1{equation~\ref{#1}}

\def\1{\bm{1}}

\DeclareMathAlphabet{\mathsfit}{\encodingdefault}{\sfdefault}{m}{sl}
\SetMathAlphabet{\mathsfit}{bold}{\encodingdefault}{\sfdefault}{bx}{n}

\usepackage{hyperref}
\usepackage{url}
\usepackage{appendix}
\usepackage{amsmath}
\usepackage{algorithm}
\usepackage{amssymb}
\usepackage[noend]{algpseudocode}
\usepackage{graphicx}
\usepackage{subcaption}
\usepackage[utf8]{inputenc}
\usepackage[export]{adjustbox}
\usepackage{wrapfig}

\algnewcommand{\LineComment}[1]{\State \(\triangleright\) {\tt #1}}

\title{PADAM: Closing The Generalization Gap of Adaptive Gradient Methods In Training Deep Neural Networks \\ \\ ICLR Reproducibilty Challenge 2019}

\author{Harshal Mittal \thanks{All the authors contributed equally to the reproducibilty challenge. The names are sorted in alphabetical order.} \\
Dept. of Electronics and Communication Engineering\\
Indian Institute of Technology, Roorkee\\
\texttt{hmittal@ec.iitr.ac.in} \\
\And
Kartikey Pandey \footnotemark[1]\\
Dept. of Electronics and Communication Engineering\\
Indian Institute of Technology, Roorkee\\
\texttt{kpandey@ec.iitr.ac.in} \\
\AND
Yash Kant \footnotemark[1]\\
Dept. of Electrical Engineering \\
Indian Institute of Technology, Roorkee\\
\texttt{ysh.kant@gmail.com} \\
}

\iclrfinalcopy %
\begin{document}

\maketitle

\begin{abstract}

This work is a part of ICLR Reproducibility Challenge 2019, we try to reproduce the results in the conference submission \textit{PADAM: Closing The Generalization Gap of Adaptive Gradient Methods In Training Deep Neural Networks}. Adaptive gradient methods proposed in past demonstrate a degraded generalization performance than the stochastic gradient descent (SGD) with momentum. The authors try to address this problem by designing a new optimization algorithm that bridges the gap between the space of Adaptive Gradient algorithms and SGD with momentum. With this method a new tunable hyperparameter called partially adaptive parameter \textit{p} is introduced that varies between [0, 0.5]. We build the proposed optimizer and use it to mirror the experiments performed by the authors. We review and comment on the empirical analysis performed by the authors. Finally, we also propose a future direction for further study of Padam. Our code is available at: \href{https://github.com/yashkant/Padam-Tensorflow}{https://github.com/yashkant/Padam-Tensorflow}

\end{abstract} 

\section{Introduction}

Adaptive gradient methods such as Adam (\cite{kingma2014adam}), Adagrad, Adadelta (\cite{zeiler2012adadelta}), RMSProp (\cite{hinton2012}), Nadam (\cite{dozat2016}), AdamW (\cite{loshchilov2017fixing}), were proposed over SGD with momentum for solving optimization of stochastic objectives in high-dimensions. Amsgrad was recently proposed as an improvement to Adam to fix convergence issues in the latter. These methods provide benefits such as faster convergence and insensitivity towards hyperparameter selection i.e. they are demonstrated to work with little tuning. On the downside, these adaptive methods have shown poor empirical performance and lesser generalization as compared to SGD-momentum. The authors of Padam attribute this phenomenon to the "over-adaptiveness" of the adaptive methods. 

The key contributions of Padam as mentioned by the authors are: 

 \begin{itemize}
 
\item {The authors put forward that Padam unifies Adam/Amsgrad and SGD with momentum by a partially adaptive parameter and that Adam/Amsgrad can be seen as a special fully adaptive instance of Padam. They further claim that Padam resolves the “small learning rate dilemma” for adaptive gradient methods and allows for faster convergence, hence closing the gap of generalization.}

\item{The authors claim that Padam generalizes equally good as SGD-momentum and achieves fastest convergence.}

\end{itemize}

We address and comment on each of the above claims from an empirical point of view. We run additional experiments to study the effect of learning rate (and its schedule) on the optimal value of partially adaptive parameter \textit{p}. From our analysis we propose the use of a suitable schedule to vary \textit{p} as the training proceeds in order to actually enjoy the best from both the worlds.

\footnotetext[1]{Authors code is available at: https://github.com/uclaml/Padam/}

\footnotetext[2]{Implementation of Adam at: \color{blue}{\href{https://github.com/tensorflow/tensorflow/blob/r1.12/tensorflow/python/training/adam.py}{https://github.com/tensorflow/tensorflow/blob/r1.12/tensorflow/python/training/adam.py}}}

\section{Background}

Padam is inspired from two recent adaptive techniques introduced in Adam and Amsgrad, we discuss them briefly here. Adam made use of bias-corrected first and second order moments along with the gradients for weight update.

  $\mathbf{\theta_{t+1}} =\mathbf{\theta_{t}} -  \alpha_t \frac{\mathbf{m_t}}{\sqrt{\mathbf{v_t}}}$ where $\mathbf{m_t} = \beta_1\mathbf{m_{t-1}} + (1 - \beta_1)\mathbf{g_t}, \mathbf{v_t} = \beta_2\mathbf{v_{t-1}} + (1 - \beta_2)\mathbf{g_t^2}$ \hspace*{2ex} (Adam)

A convergence issue in Adam was recently uncovered and addressed by Amsgrad, where they suggest tweaking the update rule slightly to fix it. Padam makes use of this updated algorithm. 

 $\mathbf{\theta_{t+1}} =\mathbf{\theta_{t}} -  \alpha_t \frac{\mathbf{m_t}}{\sqrt{\mathbf{\hat v_t}}}$ where $\mathbf{\hat v_t}$ = max($\mathbf{\hat v_{t-1}, v_t}$) \hspace*{34ex}(Amsgrad)

\subsection{Padam Algorithm}

Padam introduces a new partially adaptive parameter \textit{p} that takes value within the range [0, 0.5]. On the extremities of this range it takes the form of SGD with momentum or AMSGrad. From Algorithm 1, when \textit{p} is set to 0.0, Padam reduces to SGD with momentum whereas setting it to 0.5 leaves us with AMSGrad optimizer.

\begin{algorithm}
    \caption{Partially adaptive momentum estimation method (Padam)}\label{Likelihood Ratio Test}
    \begin{algorithmic}
        \State \textbf{input:} initial point $\mathbf{\theta_1}$ $\in$ $\mathcal{X}$; step sizes
             $\{ \alpha_t \}$; momentum parameters $\{ \beta_{1t} \}$, $\beta_2$ ; partially adaptive\\
             parameter \textit{p} $\in$ (0, 1/2]
        \State set $\mathbf{m_0 = 0, v_0 = 0, \hat v_0 = 0}$
        \State \textbf{for} \textit{t} = 1,...,\textit{T} \textbf{do}
        \State \hspace*{2ex} $\mathbf{g_t} = \nabla f_t(\mathbf{\theta_t})$
        \State \hspace*{2ex} $\mathbf{m_t} = \beta_{1t}\mathbf{m_{t-1}} + (1 - \beta_{1t})\mathbf{g_t}$
        \State \hspace*{2ex} $\mathbf{m_t} = \mathbf{m_t}/(1-\beta_1^t)$ (Compute bias-corrected first moment estimate)
        \State \hspace*{2ex} $\mathbf{v_t} = \beta_{2}\mathbf{v_{t-1}} + (1 - \beta_2)\mathbf{g_t^2}$
        \State \hspace*{2ex} $\mathbf{v_t} = \mathbf{v_t}/(1-\beta_2^t)$ (Compute bias-corrected second raw moment estimate)
        \State \hspace*{2ex} $\mathbf{\hat v_t}$ = max($\mathbf{\hat v_{t-1}, v_t}$)
        \State \hspace*{2ex} $\mathbf{\theta_{t+1}} = \Pi_{\mathcal{X},diag(\mathbf{\hat v_t^p})}(\mathbf{\theta_{t}} - \alpha_t \cdot \mathbf{m_t}/\mathbf{\hat v_t^p})$
        \State 
        \textbf{end for}
    \end{algorithmic}
\end{algorithm}

\section{Experiments}
\label{others}

In this section we describe our experiment settings for evaluating Padam. We have tried to keep our implementation faithful to the authors code \footnotemark[1]. We build three different CNN architectures as proposed in the paper, and compare Padam's performance against the other baseline algorithms. We built the Amsgrad and Padam optimizers on top of the base code of Adam in tensorflow\footnotemark[2]. 

\clearpage

\subsection{Environmental Setup}

We have built the experiments using Tensorflow version 1.13.0 and graph-free Eager Execution mode within Python 3.5.2. We ran the experiments on 4 Tesla Xp GPU cards (with 12Gb RAM per GPU).

\subsection{Datasets}
The experiments were conducted on two popular datasets for image classification: CIFAR-10  and CIFAR-100 (\cite{Krizhevsky2009LearningML}). The performance of various optimizers on the aforementioned datasets was evaluated with three different CNN architecture: VGGNet (\cite{simonyan2014deep}), ResNet (\cite{He_2016}) and Wide ResNet(\cite{zagoruyko2016wide}). We run CIFAR-10 and and CIFAR-100 task for 200 epochs. The experiments for CIFAR datasets were performed with a learning rate decay at every 50th epoch ie. (50,100,150). We were unable to perform the experiments on the ImageNet dataset because of the time constraints and limited availability of computing resources.  

\subsection{Baseline Algorithms}

We compare Padam against the most popular adaptive gradient optimizers and SGD-momentum. Note that the evaluation against AdamW was added by the authors at a later stage and the details about it were not completely disclosed in the updated version of the paper or code, owing to this delay we have not been able to carry out experiments with AdamW.

\begin{table}[h]
\caption{Baseline Algorithms and their corresponding hyperparameters}
\label{sample-table}
\begin{center}
\begin{tabular}{l|ccccc}
\hline
\multicolumn{1}{l}{\bf Optimizer}  &\multicolumn{1}{c}{\bf Padam \footnotemark[3]} &\multicolumn{1}{c}{\bf SGD+Momentum} &\multicolumn{1}{c}{\bf Adam} &\multicolumn{1}{c}{\bf Amsgrad}
\\ \hline
Initial Learning rate &0.1 &0.1 &0.001 &0.001 \\
Beta1 &0.9 &- &0.9 &0.9 \\
Beta2 &0.999 &- &0.99 &0.99 \\
Weight decay &0.0005 &0.0005 &0.0001 &0.0001 \\
Momentum &- &0.9 &- &- \\
\hline
\end{tabular}
\end{center}
\end{table}
\footnotetext[3]{We have used p=0.125 as the hyperparameter for the experiments unless specifically mentioned}
\begin{figure}[!]
 
\begin{subfigure}{0.2\textwidth}
\includegraphics[width=0.9\linewidth, height=20cm]{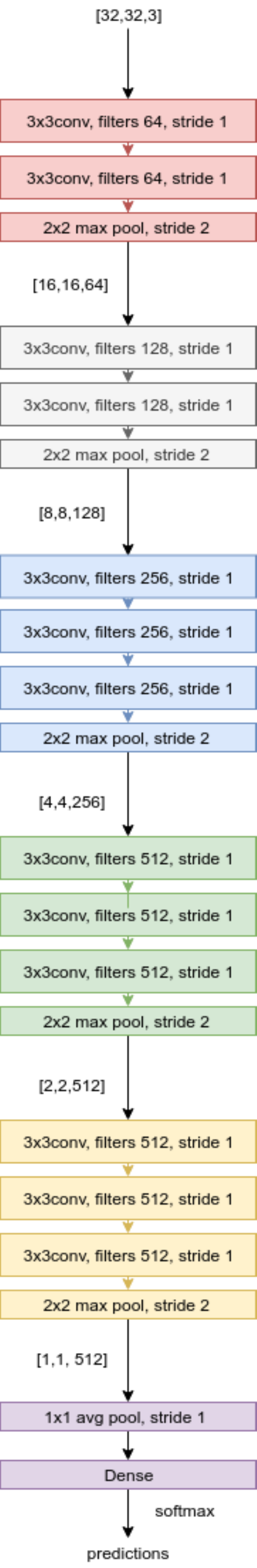}
\caption{VGG-Net}
\label{fig:subim1}
\end{subfigure}
\begin{subfigure}{0.4\textwidth}
\hspace*{0.75cm}
\includegraphics[width=0.9\linewidth, height=20cm]{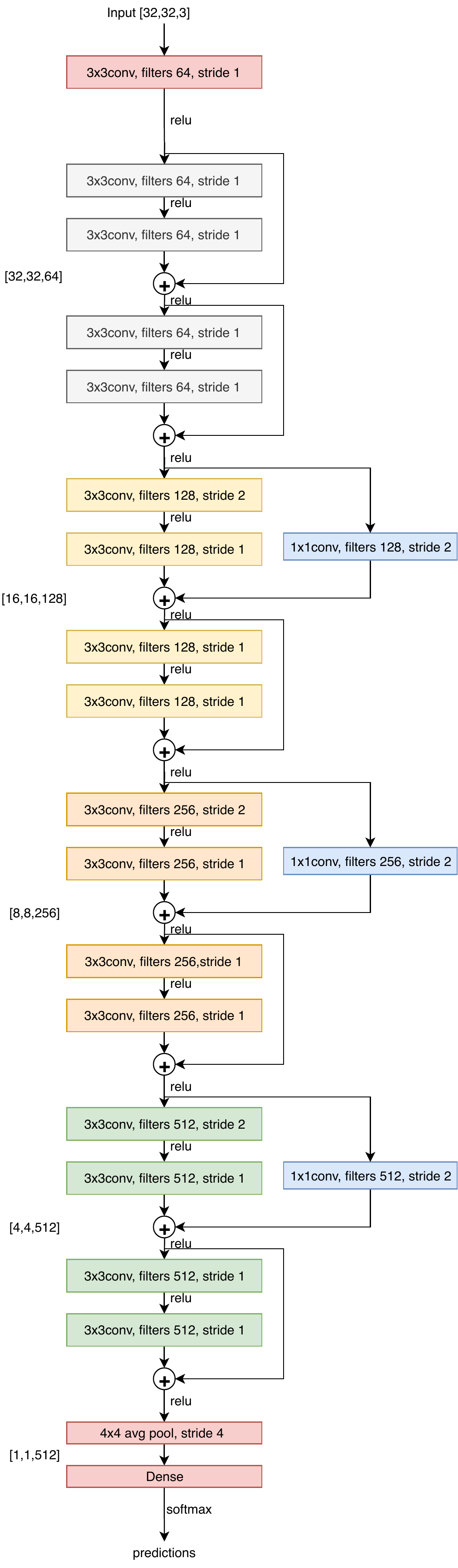}
\caption{ResNet}
\label{fig:subim2}
\end{subfigure}
\begin{subfigure}{0.4\textwidth}
\hspace*{0.5cm}
\includegraphics[width=0.9\linewidth, height=20cm]{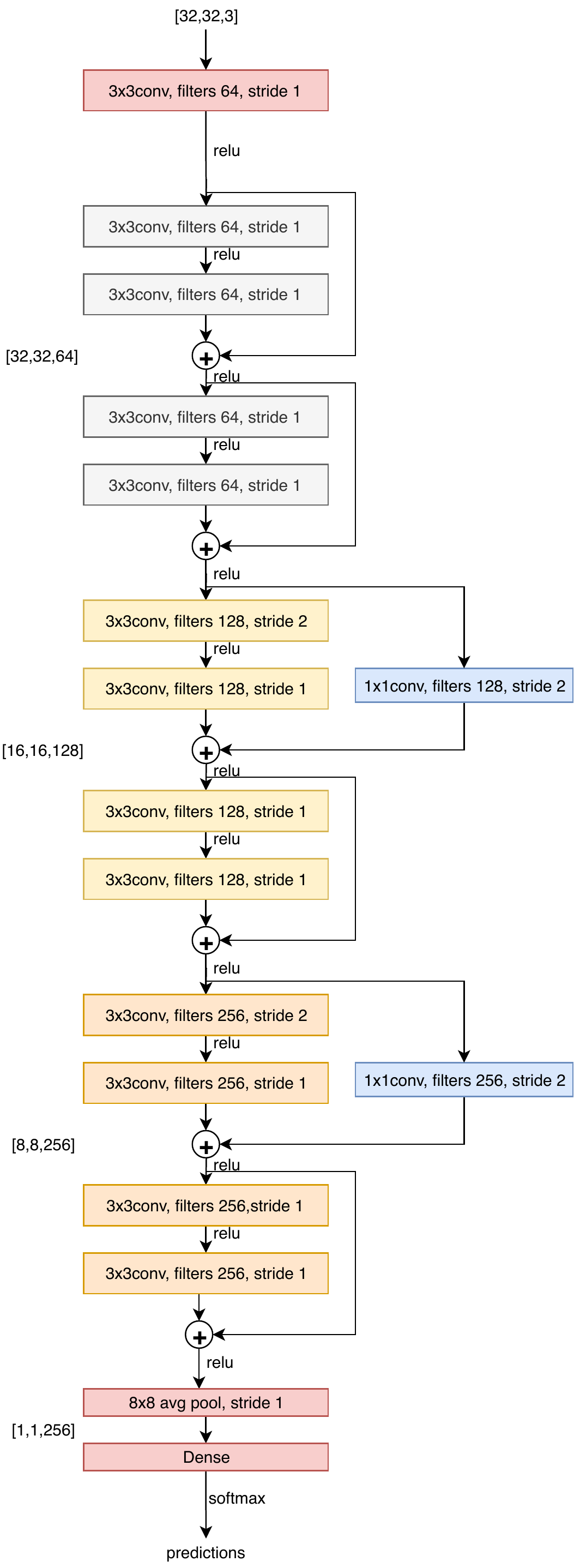}
\caption{WideResNet}
\label{fig:subim3}
\end{subfigure}
\caption{Architectures used in experiments. Note that we do not explicitly show batch normalization layer after each convolution operation for brevity.}
\label{fig:architechture}
\end{figure}

\subsection{Architectures}

We have built architectures faithful to the code released by the authors, and are shown in Figure 1.

\subsubsection{VGGNet}
The VGG-16 network uses only 3 x 3 convolutional layers stacked on top of each other for increasing depth and adopts max pooling to reduce volume size. Finally, two fully-connected layers are followed by a softmax classifier.

\subsubsection{ResNet}
Residual Neural Network (ResNet) \cite{He_2016} introduces a novel architecture with “skip connections” and features a heavy use of batch normalization. As per authors, we use ResNet-18 for this experiment, which contains 4 blocks each comprising of 2 basic building blocks.

\subsubsection{Wide ResNet}
Wide Residual Network \cite{zagoruyko2016wide} further exploits the “skip connections” used in ResNet and in the meanwhile increases the width of residual networks. In detail, we use the 16 layer Wide ResNet with 4 multipliers (WRN-16-4) in the experiments.

\section{Evaluation and Results}

In this section we comment on the results we obtained with this reproducibility effort. We divide this section into four parts.

\subsection{Train Experiments}

Padam is compared with other proposed baselines. Figure 2 demonstrates the Top-5 Test Error and Figure 3 shows the Train Loss and Test Error for the three architectures on CIFAR-10. We find that Padam performs comparably with SGD-momentum on all the three architectures in test error and maintains a rate of convergence between Adam/AMSgrad and SGD. Padam works as proposed by the authors to bridge the gap between adaptive methods and SGD at the cost of introducing a new hyperparameter \textit{p}, which requires tuning. Although, we don't see a clear motivation behind the grid-search approach used by the authors to select the value of this partially adaptive parameter \textit{p}.

The results on CIFAR-100 can be found in the appendix.

\begin{figure}[h]
\begin{subfigure}{0.5\textwidth}
\includegraphics[width=1\linewidth, height=5cm]{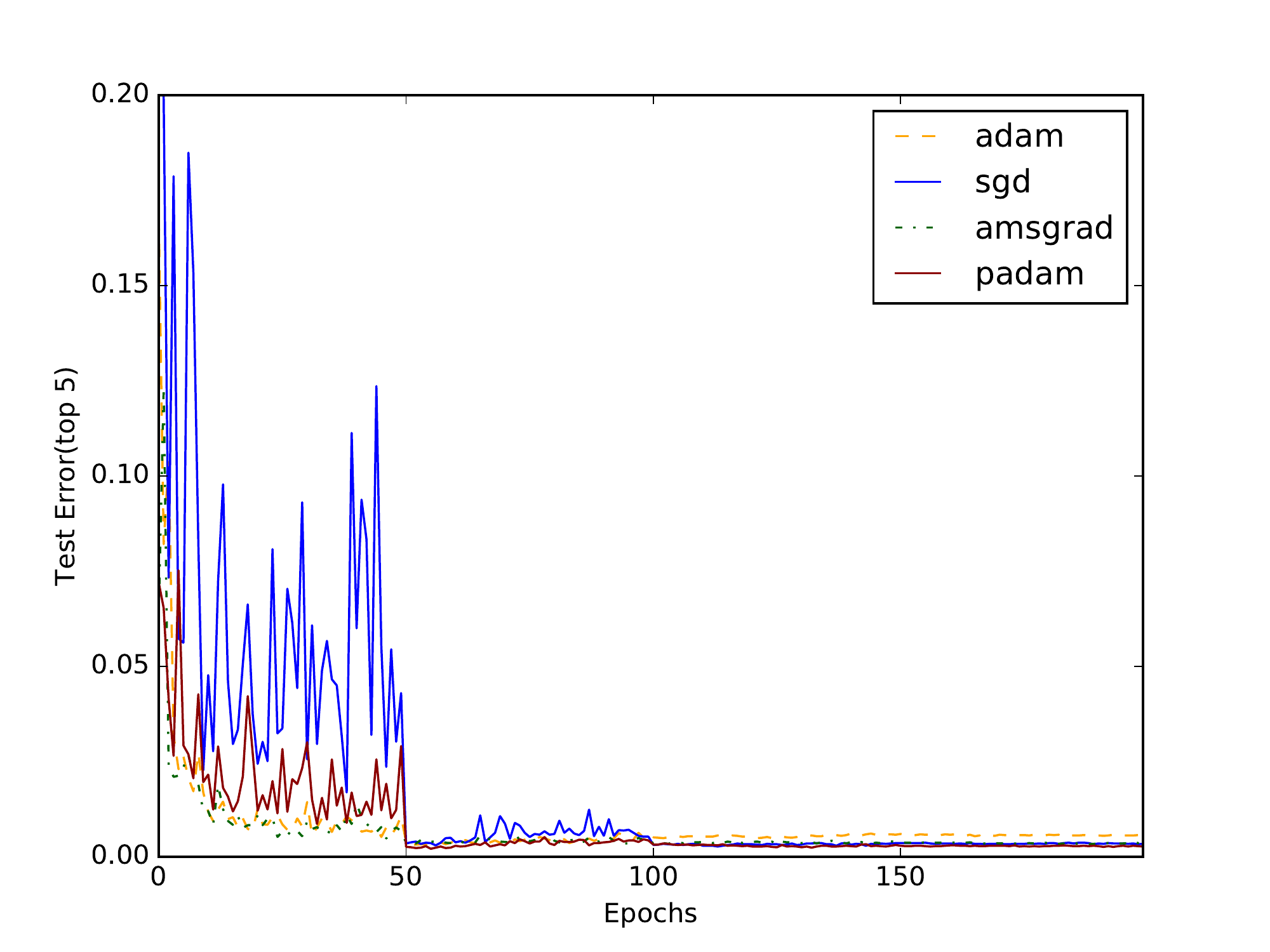}
\caption{Top-5 Error for VGGNet}
\label{fig:subim4}
\end{subfigure}
\begin{subfigure}{0.5\textwidth}
\includegraphics[width=1\linewidth, height=5cm]{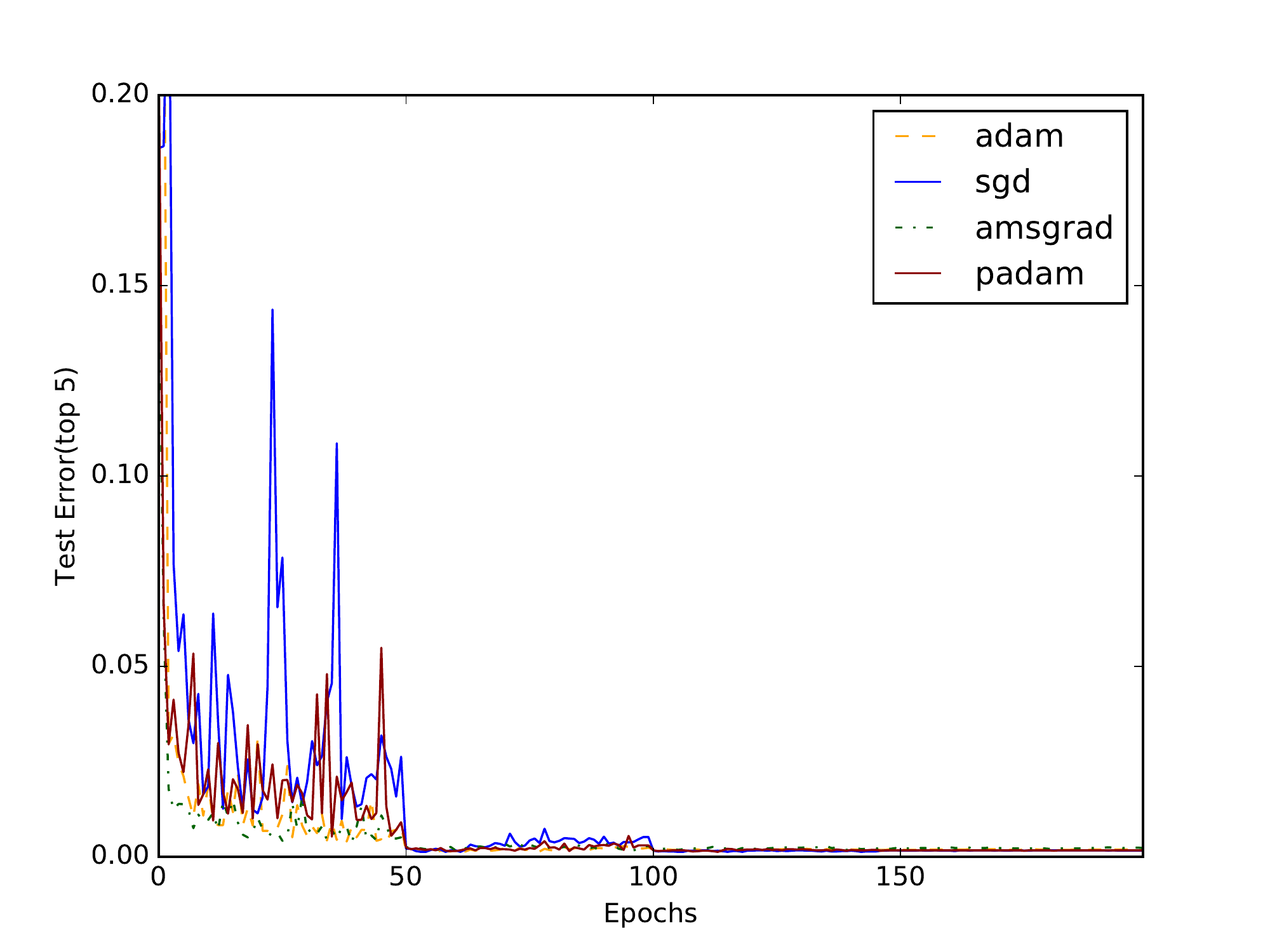}
\caption{Top-5 Error for ResNet}
\label{fig:subim5}
\end{subfigure}
\begin{center}
\begin{subfigure}{0.5\textwidth}
\includegraphics[width=1\linewidth, height=5cm]{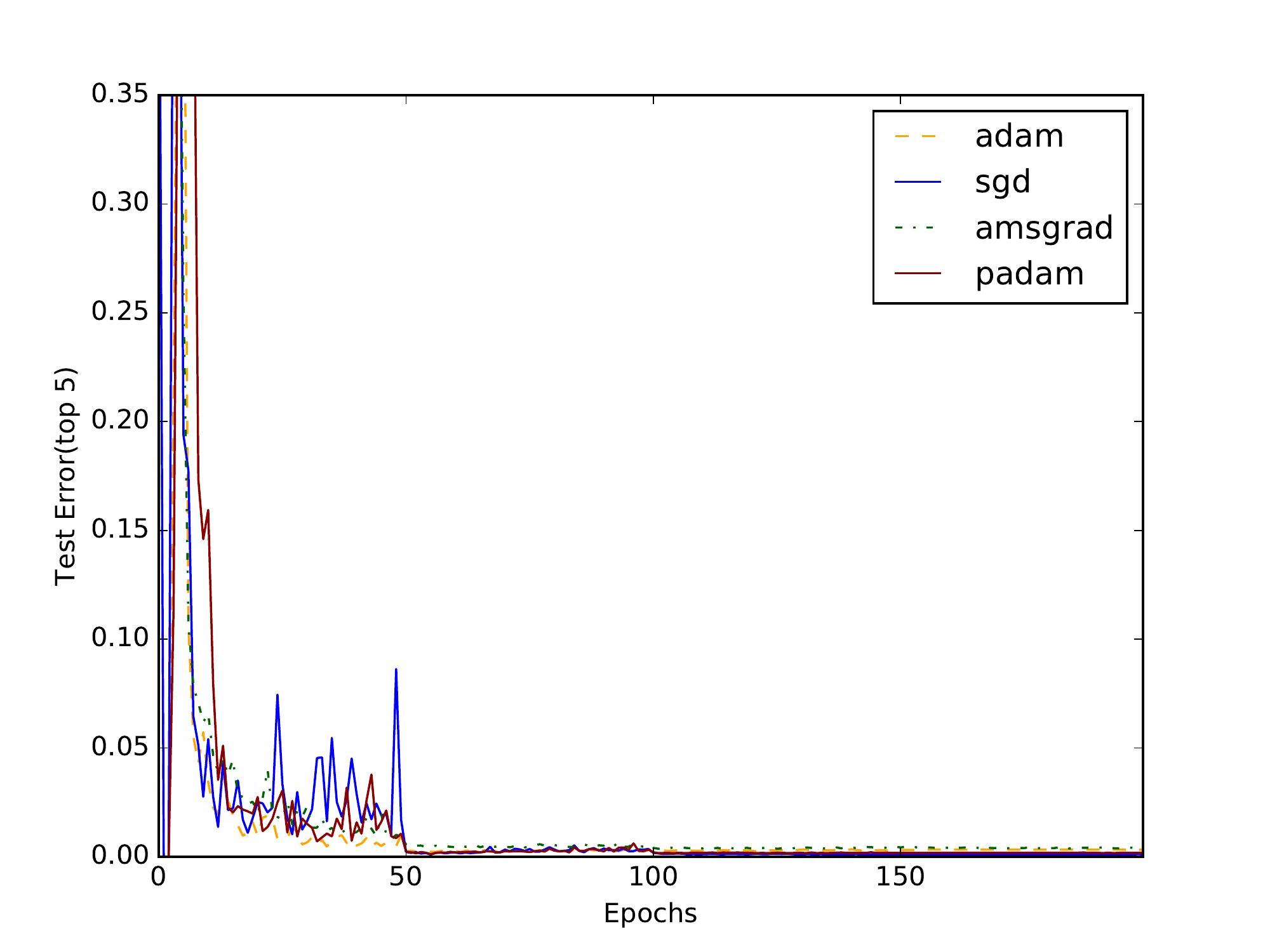}
\caption{Top-5 Error for Wide ResNet}
\label{fig:subim6}
\end{subfigure}
\end{center}
\caption{Top-5 error for three CNN architectures on CIFAR-10.}
\label{fig:image21}
\end{figure}

\clearpage

\begin{figure}[h]
 
\begin{subfigure}{0.5\textwidth}
\includegraphics[width=1\linewidth, height=5cm]{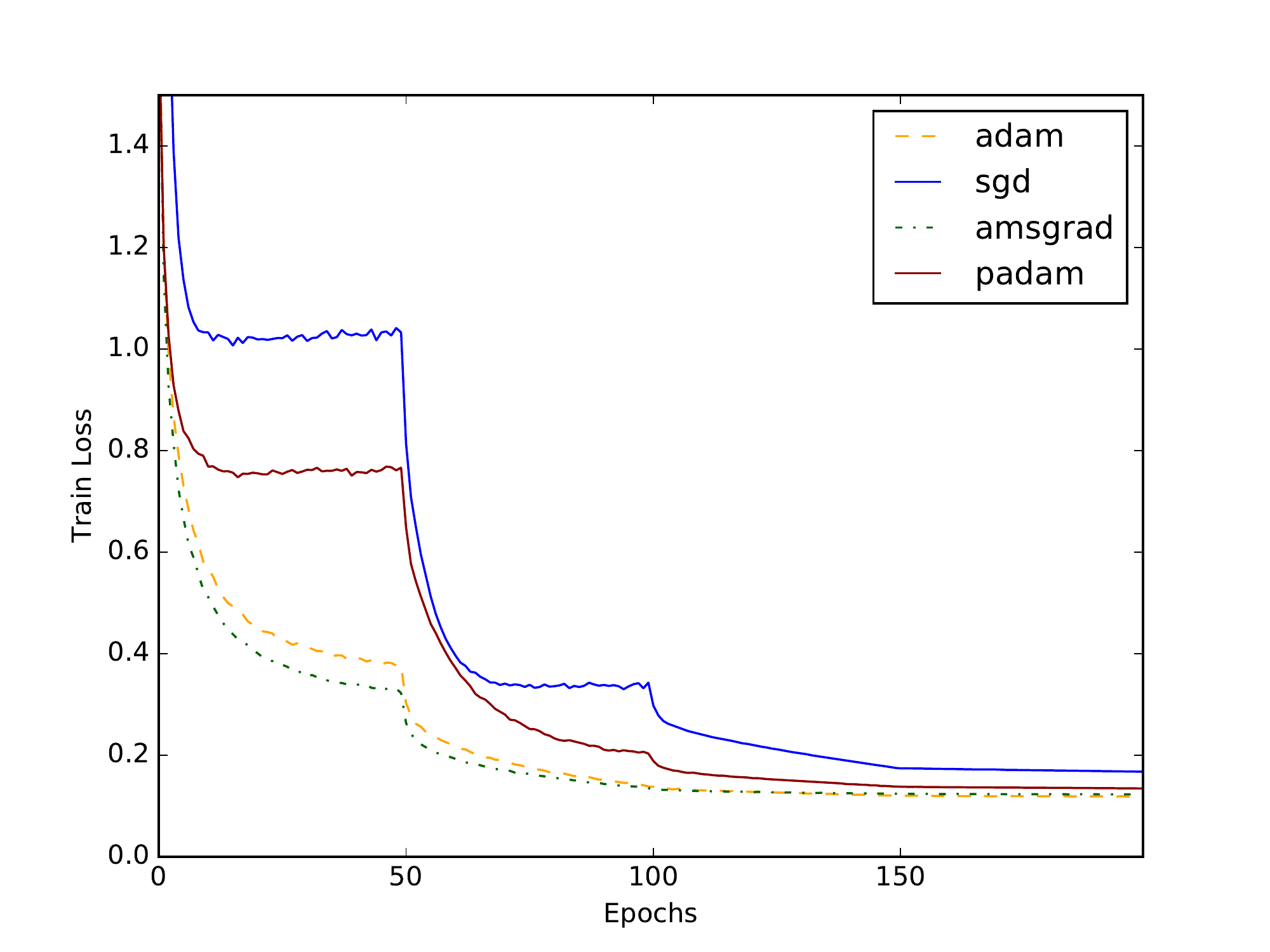}
\caption{Train Loss for VGGNet}
\label{fig:subim7}
\end{subfigure}
\begin{subfigure}{0.5\textwidth}
\includegraphics[width=1\linewidth, height=5cm]{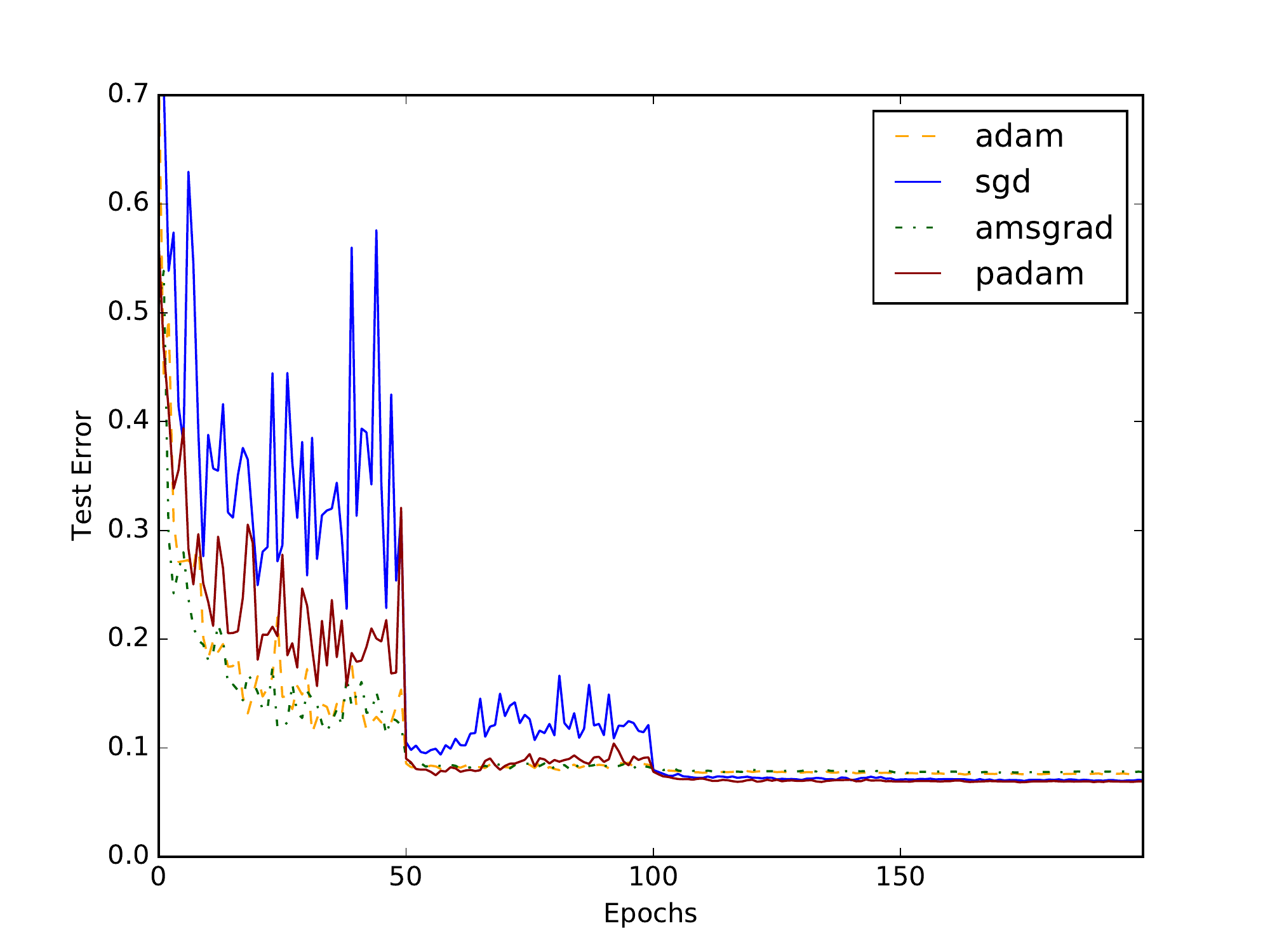}
\caption{Test Error for VGGNet}
\label{fig:subim8}
\end{subfigure}
\begin{subfigure}{0.5\textwidth}
\includegraphics[width=1\linewidth, height=5cm]{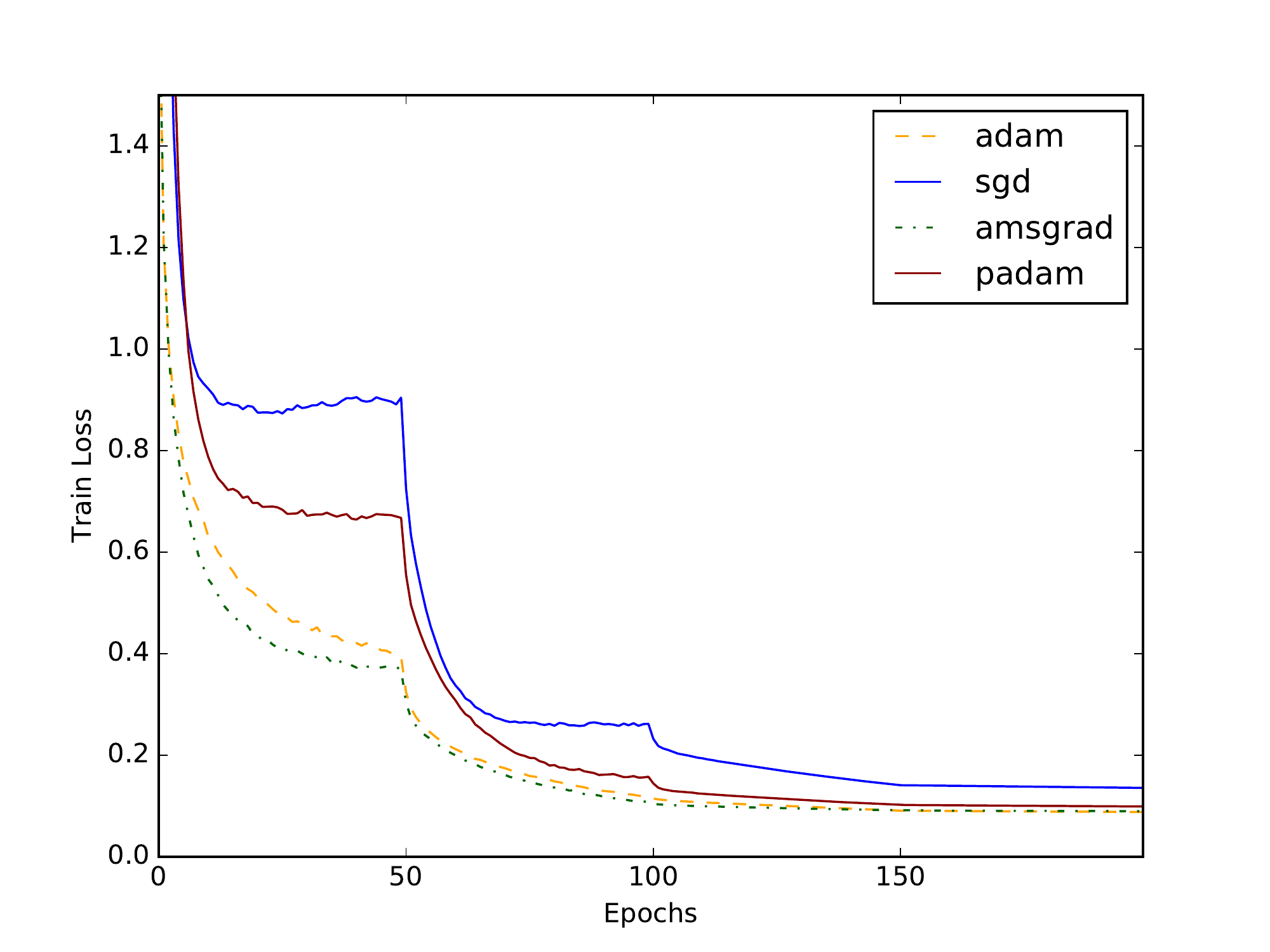}
\caption{Train Loss for ResNet}
\label{fig:subim9}
\end{subfigure}
\begin{subfigure}{0.5\textwidth}
\includegraphics[width=1\linewidth, height=5cm]{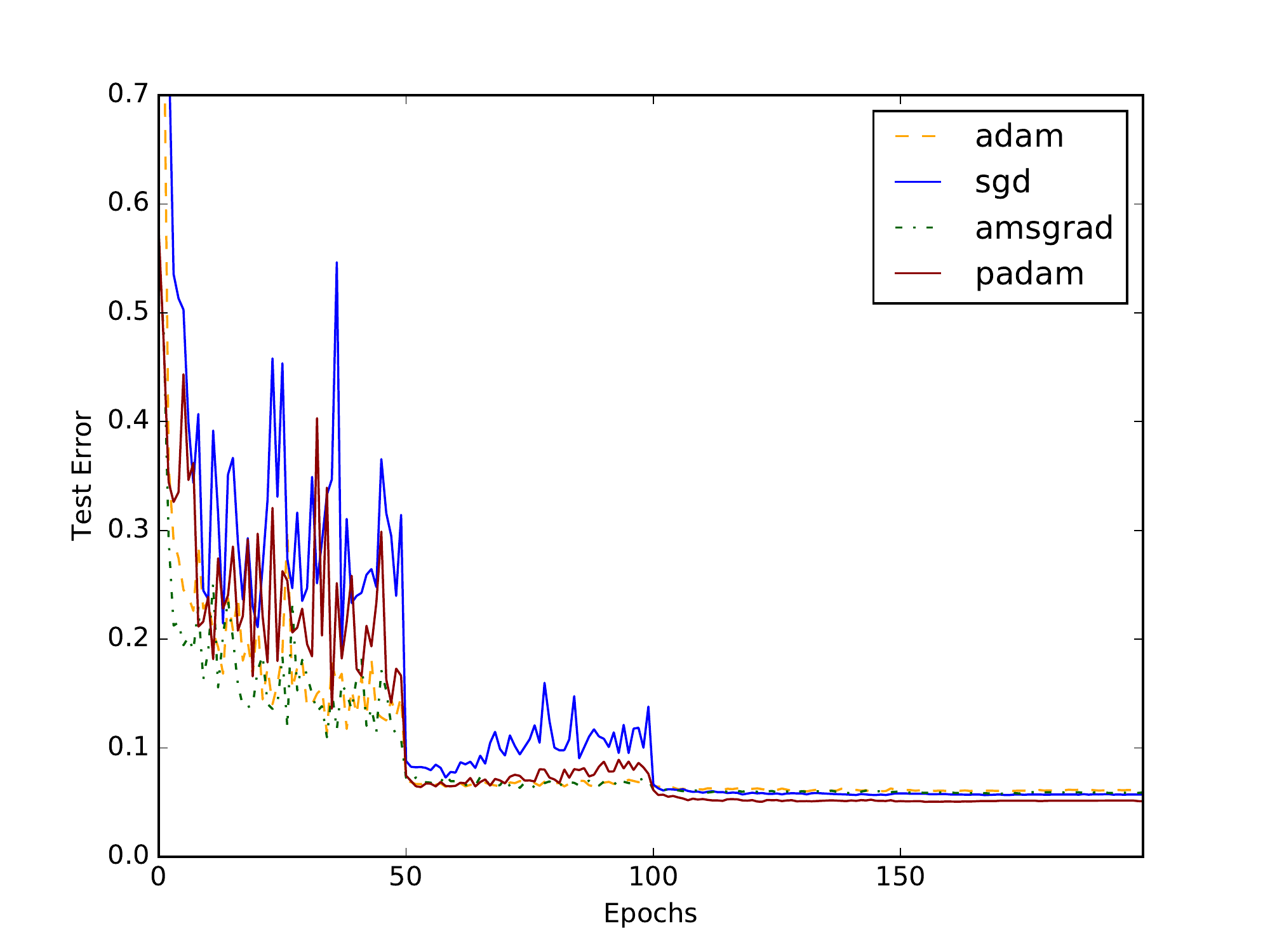}
\caption{Test Error for ResNet}
\label{fig:subim10}
\end{subfigure}
\begin{subfigure}{0.5\textwidth}
\includegraphics[width=1\linewidth, height=5cm]{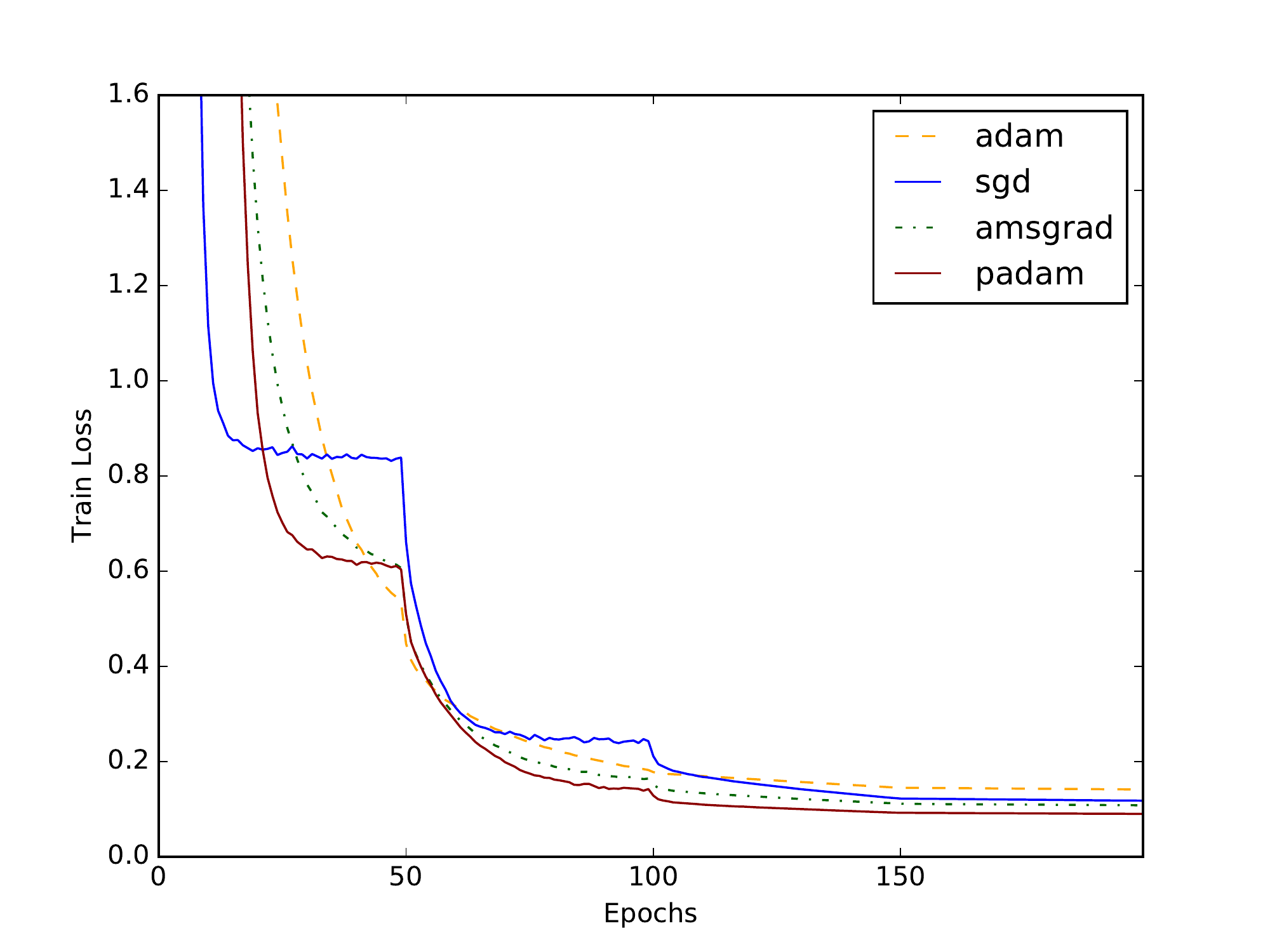}
\caption{Train Loss for Wide ResNet}
\label{fig:subim11}
\end{subfigure}
\begin{subfigure}{0.5\textwidth}
\includegraphics[width=1\linewidth, height=5cm]{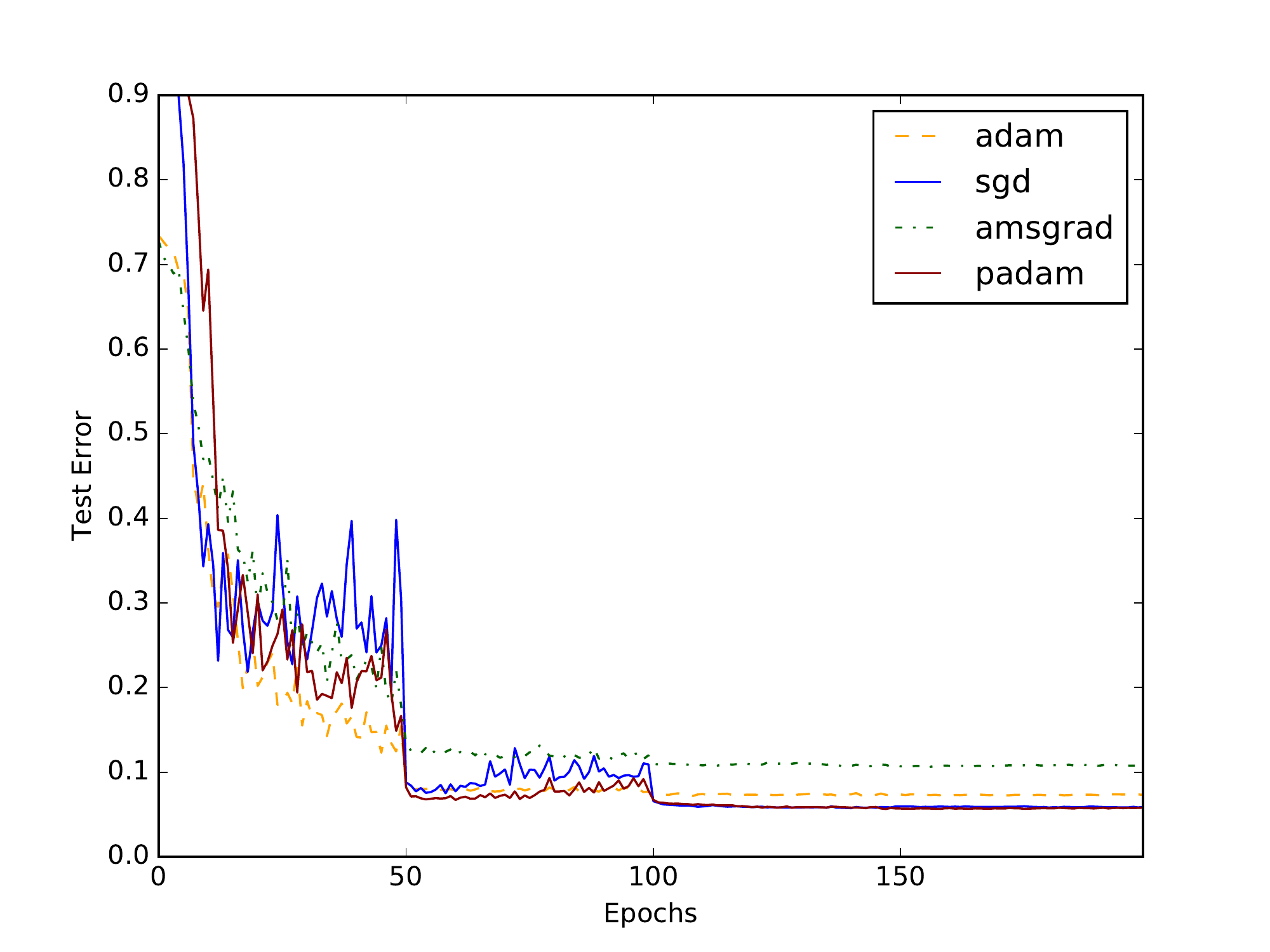}
\caption{Test Error for Wide ResNet}
\label{fig:subim12}
\end{subfigure}
\caption{Train loss and test error (top-1 error) of three CNN architectures on CIFAR-10.}
\label{fig:image22}
\end{figure}

\begin{table}[H]
\caption{Test Accuracy of VGGNet on CIFAR-10}
\label{sample-table1}
\begin{center}
\begin{tabular}{l|ccccc}
 \hline 
\multicolumn{1}{l}{\bf Methods}  &\multicolumn{1}{c}{\bf 50th epoch} &\multicolumn{1}{c}{\bf 100th epoch} &\multicolumn{1}{c}{\bf 150th epoch} &\multicolumn{1}{c}{\bf 200th epoch}
\\ \hline 
SGD Momentum &68.71 &87.88 &92.94 &92.95 \\
Adam &84.62 &91.54 &92.34 &92.39 \\
Amsgrad &{\bf 87.89} &{\bf 91.75}  &92.26 &92.19 \\
Padam &67.92 &90.86 &{\bf 93.08} &{\bf 93.06} \\ \hline
\end{tabular}
\end{center}
\end{table}

\subsection{\textit{p}-value Experiments}
In order to find an optimal working value of \textit{p} the authors perform grid search over three options [0.25, 0.125, 0.0625]. They do so by keeping the base learning rate fixed to 0.1 and decaying it by a factor of 0.1 per 30 epochs. We perform the same experiment on CIFAR-10 and CIFAR-100, the results are plotted in Figure 4. We observe similar results as the authors, and find the most optimal setting for \textit{p} to be 0.125, out of the proposed three values. Nevertheless, we would like to press that this value of \textit{p} is sub-optimal and may turn out to be sensitive to the learning rate's base value. To analyze this we perform sensitivity experiments of \textit{p} against various learning rates and it turns out that \textit{p} is indeed sensitive to it.

\begin{figure}[h]
\begin{subfigure}{0.5\textwidth}
\includegraphics[width=1\linewidth, height=5cm]{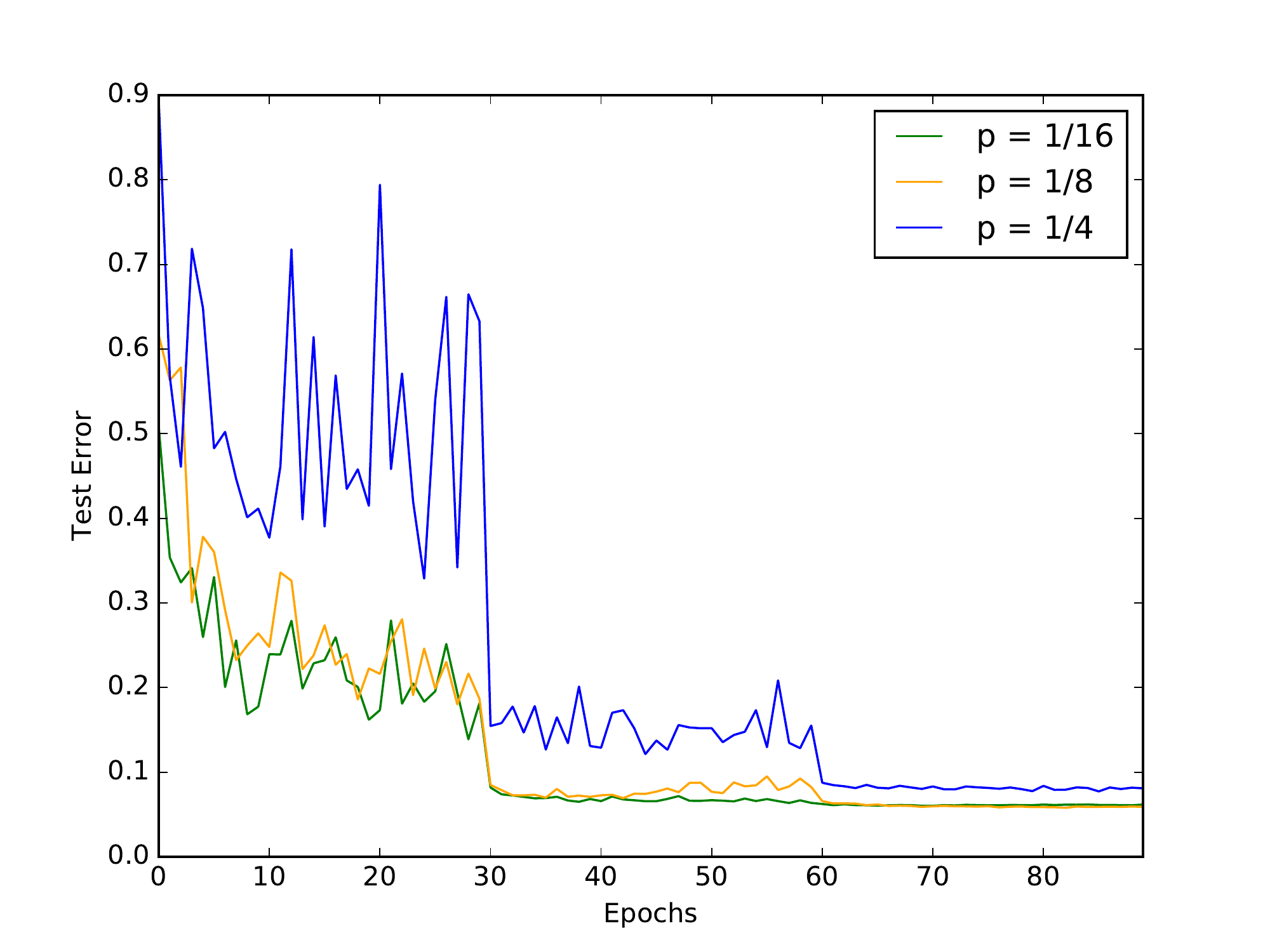}
\caption{CIFAR-10}
\label{fig:subim13}
\end{subfigure}
\begin{subfigure}{0.5\textwidth}
\includegraphics[width=1\linewidth, height=5cm]{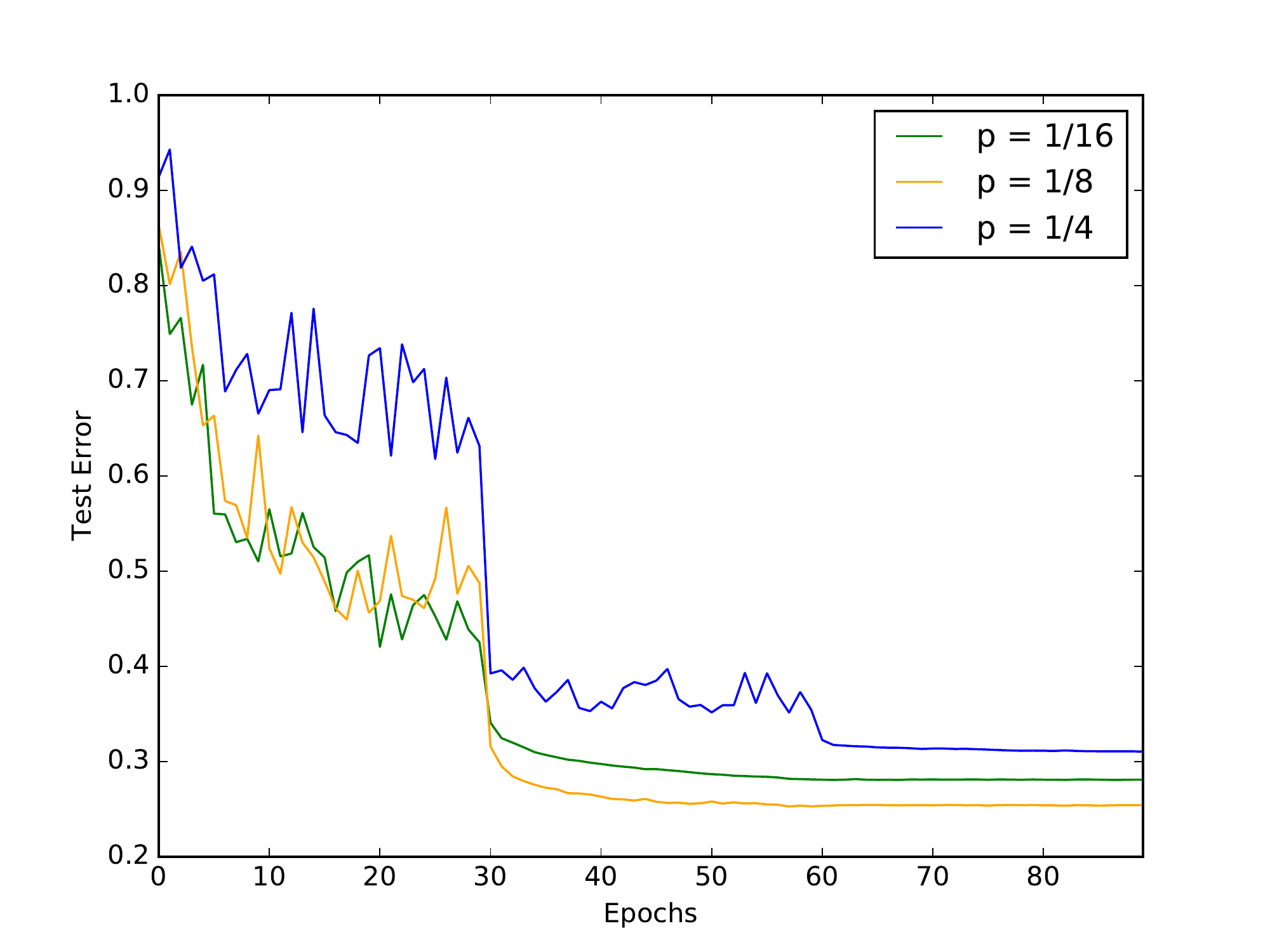}
\caption{CIFAR-100}
\label{fig:subim14}
\end{subfigure}
\caption{Performance comparison of Padam with different choices of p for training ResNet on CIFAR-10 / CIFAR-100 dataset.}
\label{fig:image23}
\end{figure}

\subsection{Sensitivity Experiments}
To evaluate the possibility that optimal value of partial adaptive parameter \textit{p} is entangled with learning rate we run the sensitivity experiments. We perform experiments with three fixed values of \textit{p} from {0.25, 0.125, 0.0625}. For each fixed value of \textit{p} we vary the base learning rate over \{0.1, 0.01, 0.001\}. We run each evaluation for 30 epochs on CIFAR-10 and CIFAR-100 with ResNet. We expect that this would uncover the dependence of \textit{p} on base learning rate.

The results for CIFAR-10 are plotted in Figure 5. From  Figure 5(b) we observe that with p = 0.25 base learning rate of 0.01 or 0.001 seems to be a more appropriate choice as compared to 0.1 owing to its better Test Error performance. As we decrease the value of p, we find that higher base learning rates start performing better as evident from Figure 5(d) and 5(f). This observation favors the argument that \textit{p} is indeed sensitive to the base learning rate.

Results of sensitivity experiments on CIFAR-100 are moved to Appendix.  

\subsection{Proposed Further Study}

From the sensitivity experiments we can infer that while using higher values of \textit{p} Padam behaves more adaptive-like (performs better with lower learning rates) and with smaller values of \textit{p} Padam demonstrates a behavior closer to SGD (performs better with higher learning rates).

Our primary objective behind designing Padam was to achieve two things: good convergence (initially) and better generalization (finally). In order to do so we would like Padam to behave adaptive-like initially and SGD-like finally. In this way Padam would be able to exploit both worlds to their fullest within the training life-cycle. 

To do so we propose to initialize Padam with high \textit{p} and low base learning rate and then decay \textit{p} during the training life-cycle. Correspondingly the learning rate can be mildly decreased in the middle or towards the end of the training cycle in order to  generate conditions for the SGD-like Padam to converge.  

Recently, AdamW has demonstrated better generalization by decoupling the mechanism of weight decay from the update rule, this method can also further compliment Padam's result. We haven't been able to finish running these proposed experiments due to time and resource constraints.

\begin{figure}[h]
 
\begin{subfigure}{0.5\textwidth}
\includegraphics[width=1\linewidth, height=5cm]{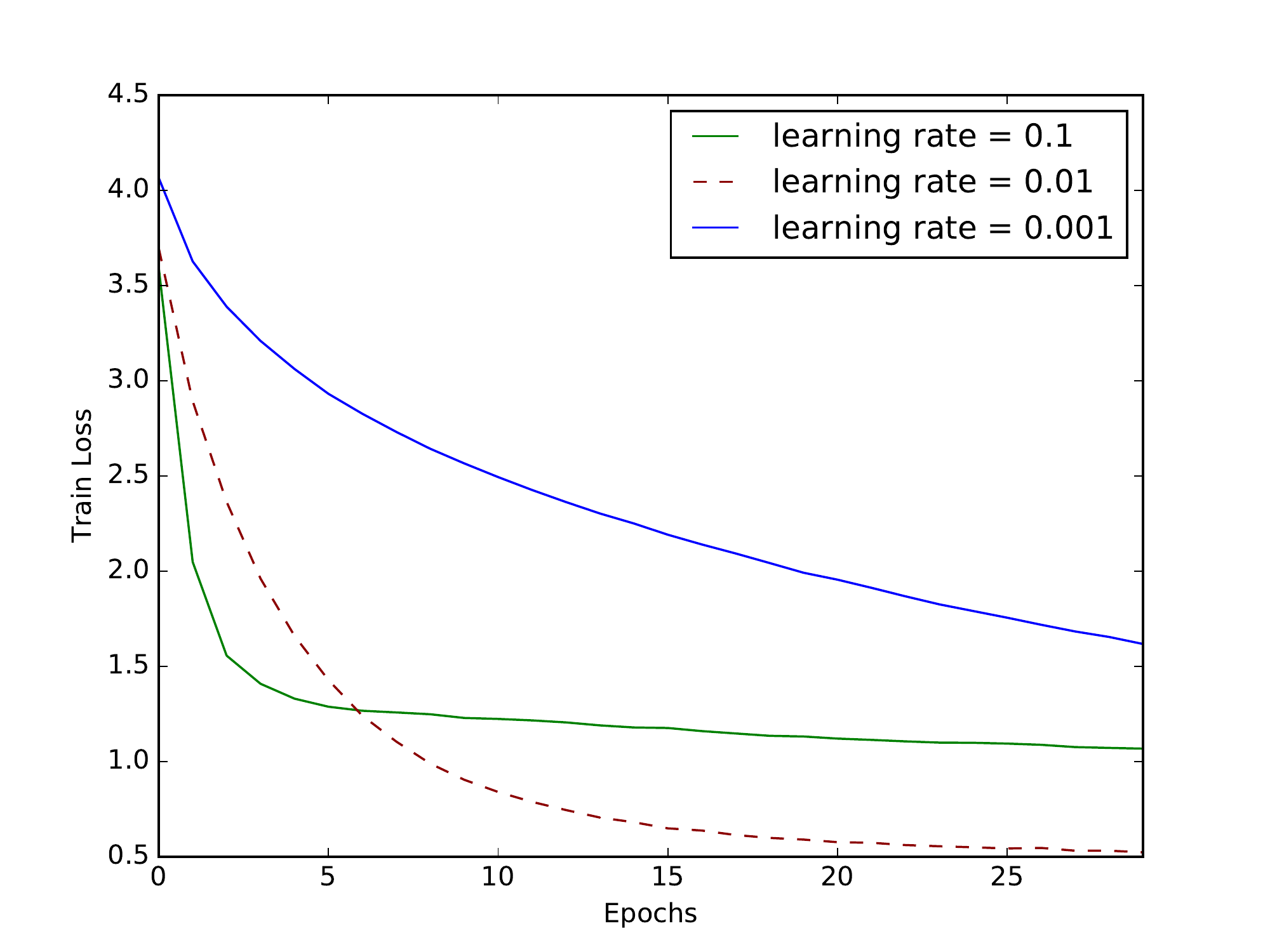}
\caption{Train Loss with p = 0.25}
\label{fig:subim15}
\end{subfigure}
\begin{subfigure}{0.5\textwidth}
\includegraphics[width=1\linewidth,height=5cm]{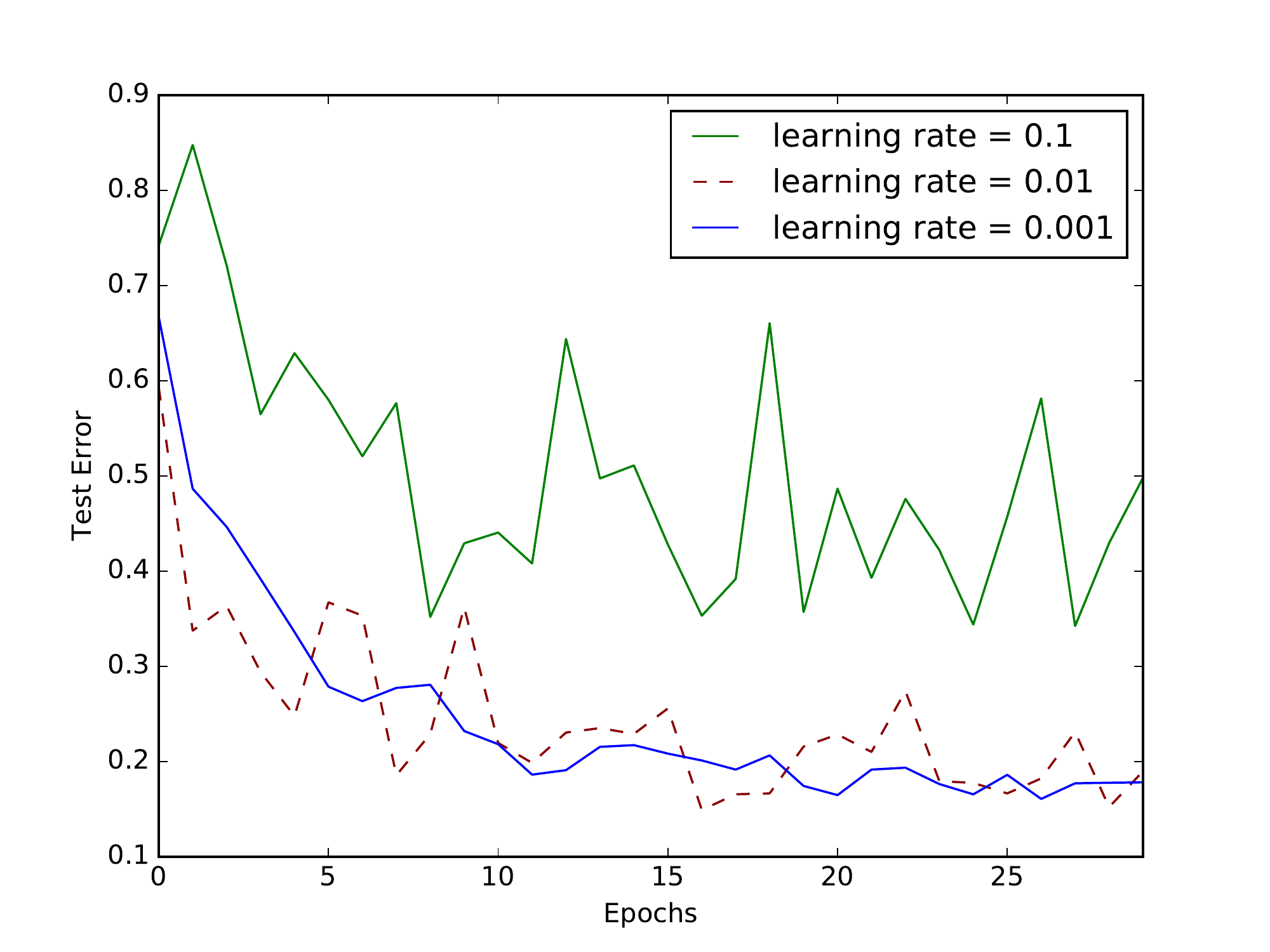}
\caption{Test Error with p = 0.25}
\label{fig:subim16}
\end{subfigure}
\begin{subfigure}{0.5\textwidth}
\includegraphics[width=1\linewidth, height=5cm]{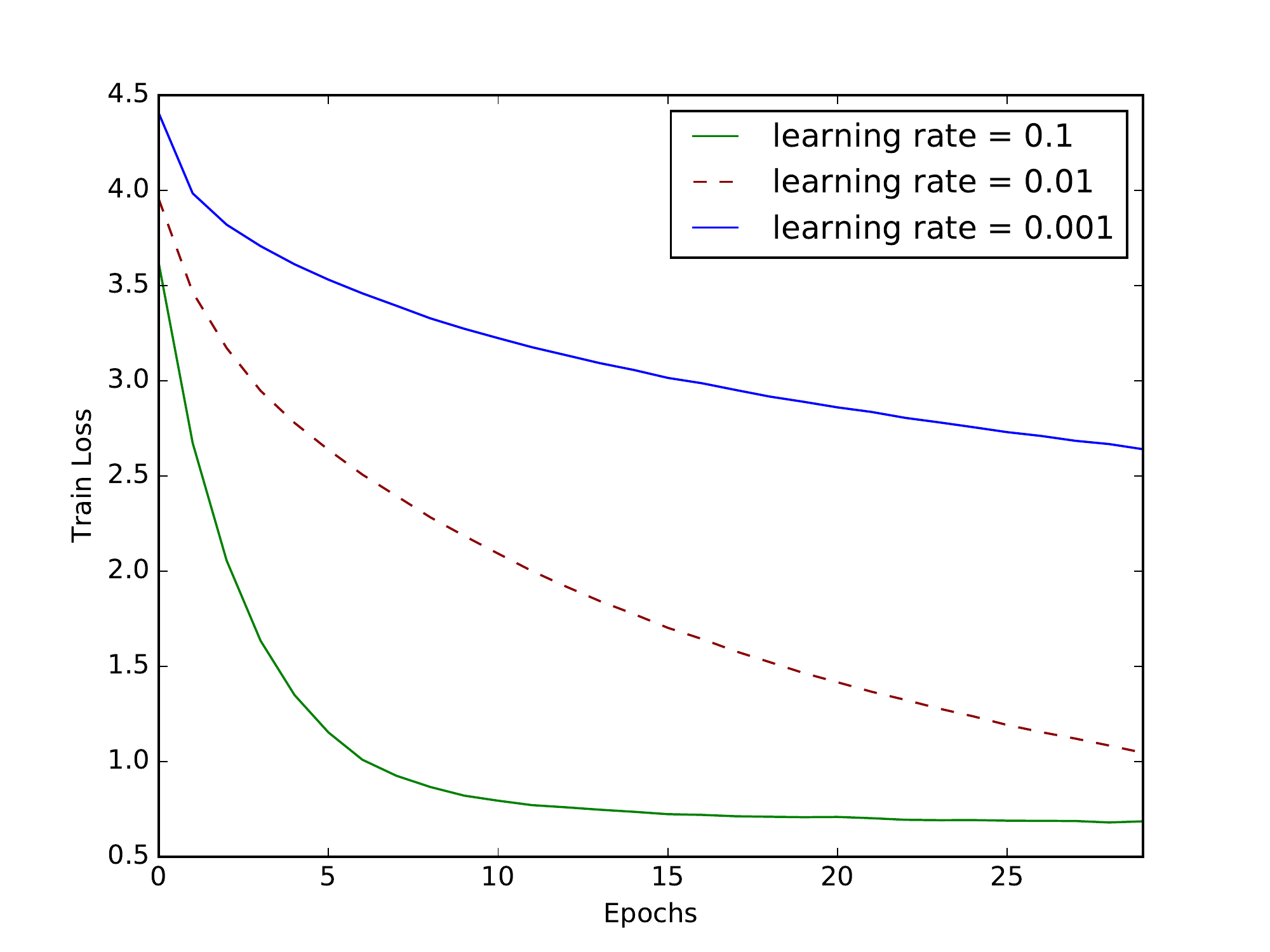}
\caption{Train Loss with p = 0.125}
\label{fig:subim17}
\end{subfigure}
\begin{subfigure}{0.5\textwidth}
\includegraphics[width=1\linewidth, height=5cm]{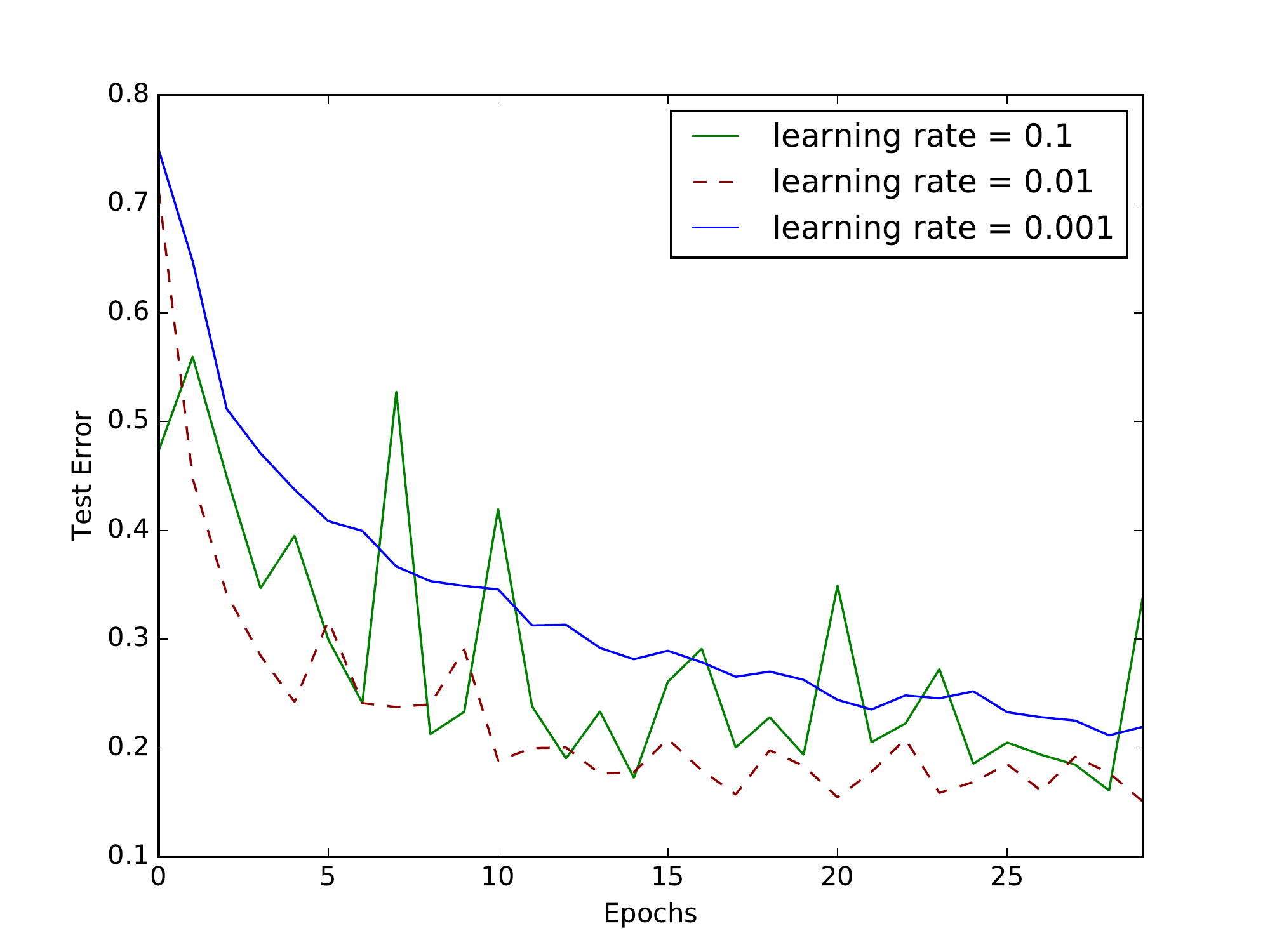}
\caption{Test Error with p = 0.125}
\label{fig:subim18}
\end{subfigure}
\begin{subfigure}{0.5\textwidth}
\includegraphics[width=1\linewidth, height=5cm]{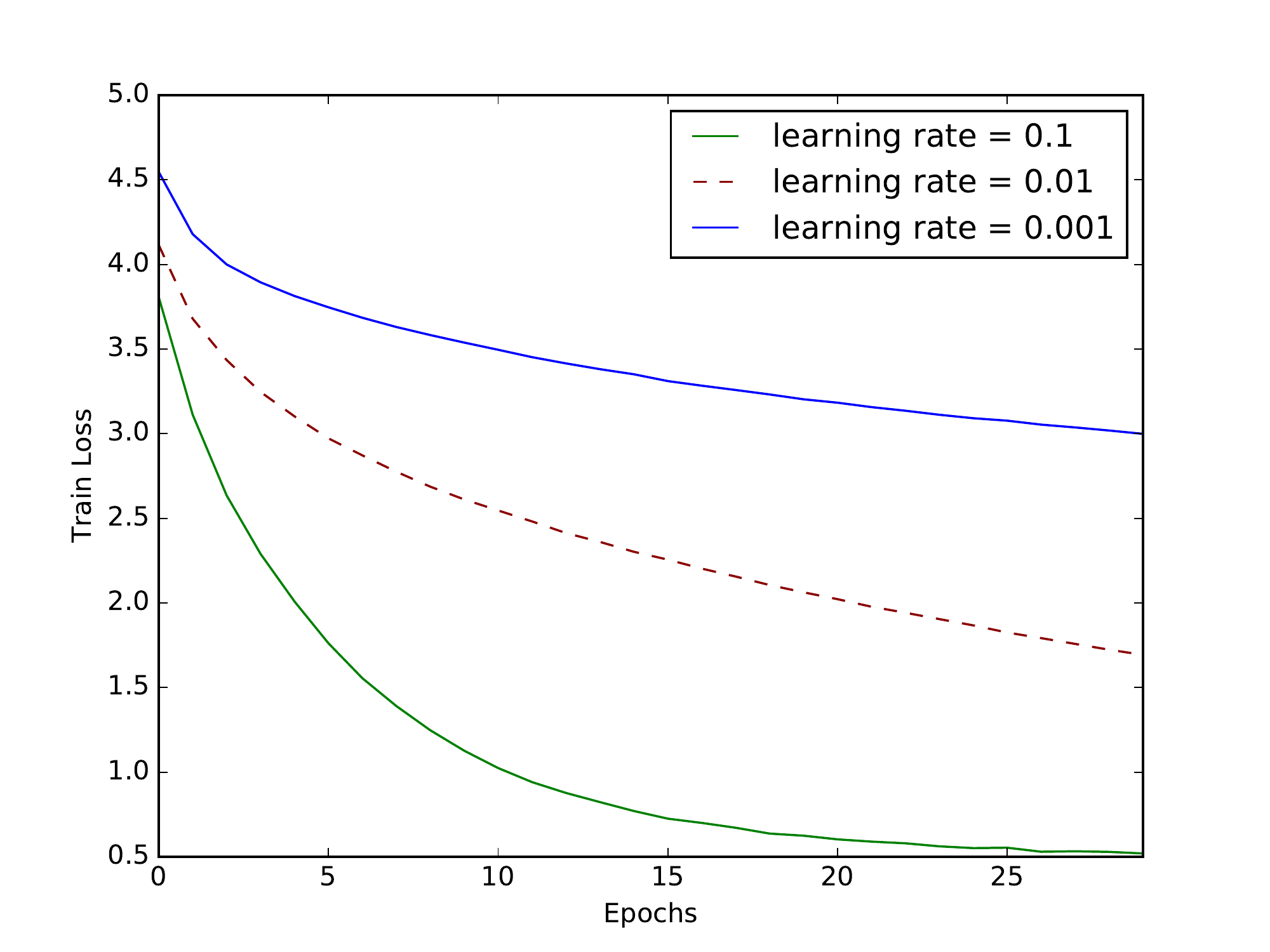}
\caption{Train Loss with p = 0.0625}
\label{fig:subim19}
\end{subfigure}
\begin{subfigure}{0.5\textwidth}
\includegraphics[width=1\linewidth, height=5cm]{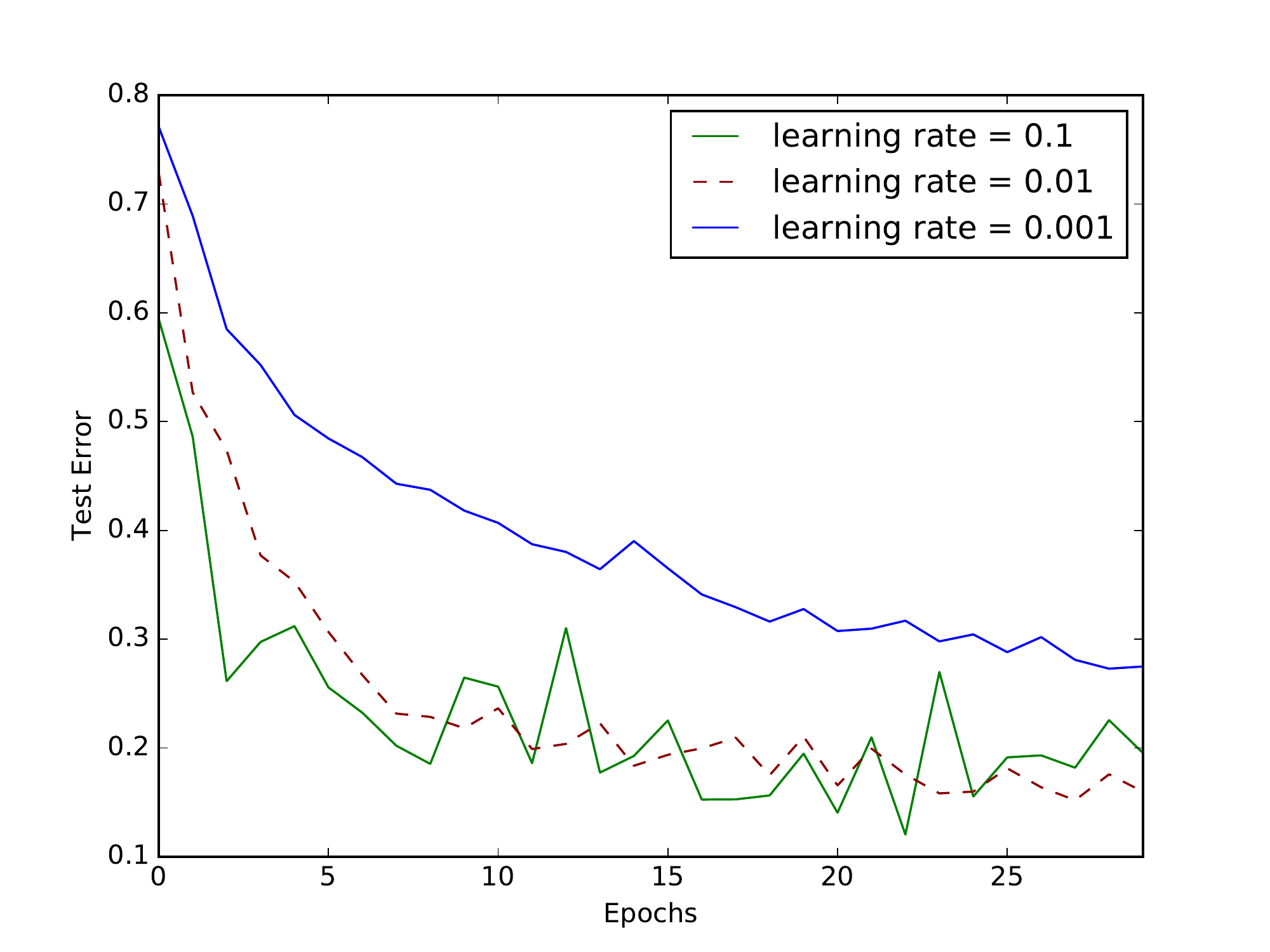}
\caption{Test Error with p = 0.0625}
\label{fig:subim20}
\end{subfigure}
\caption{Performance comparison of Padam with different choices of learning rate for three different fixed values of p (0.25, 0.125, 0.0625) for ResNet on CIFAR-10 dataset .}
\label{fig:image4}
\end{figure}

\clearpage

\section{Discrepancies, Suggestions and Conclusion}

The authors argue that adaptive gradient methods when used with larger base learning rate gives rise to the gradient explosion problem because of the presence of the second order moment term $v$ in the denominator. This proposition implicitly assumes $v$ to be in between 0 and 1, which might not always be the case and hence the factor may cause the effective learning rate to either increase or decrease.

Overall, we conclude from the empirical evaluation that Padam is capable of mixing the benefits of adaptive gradient methods with those of SGD with momentum. Perhaps studying the newly introduced partially adaptive \textit{p} parameter seems to be a good direction to further this work along.

\bibliography{iclr2019_conference}
\bibliographystyle{iclr2019_conference}
\clearpage
\appendix

\section{Experiments on CIFAR-100}

\begin{figure}[h]
 
\begin{subfigure}{0.5\textwidth}
\includegraphics[width=1\linewidth, height=5cm]{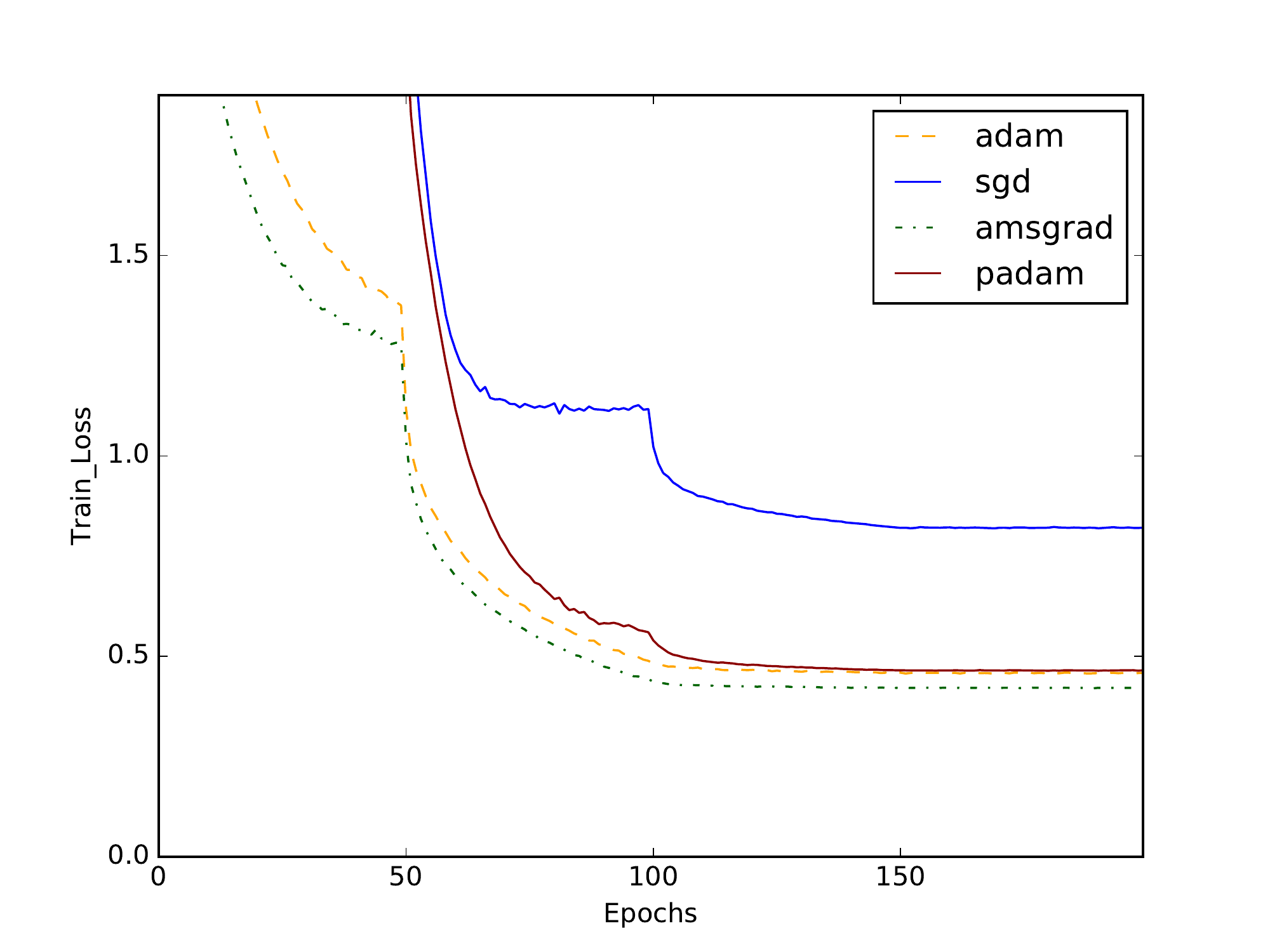}
\caption{Train Loss for VGGNet}
\label{fig:subim21}
\end{subfigure}
\begin{subfigure}{0.5\textwidth}
\includegraphics[width=1\linewidth, height=5cm]{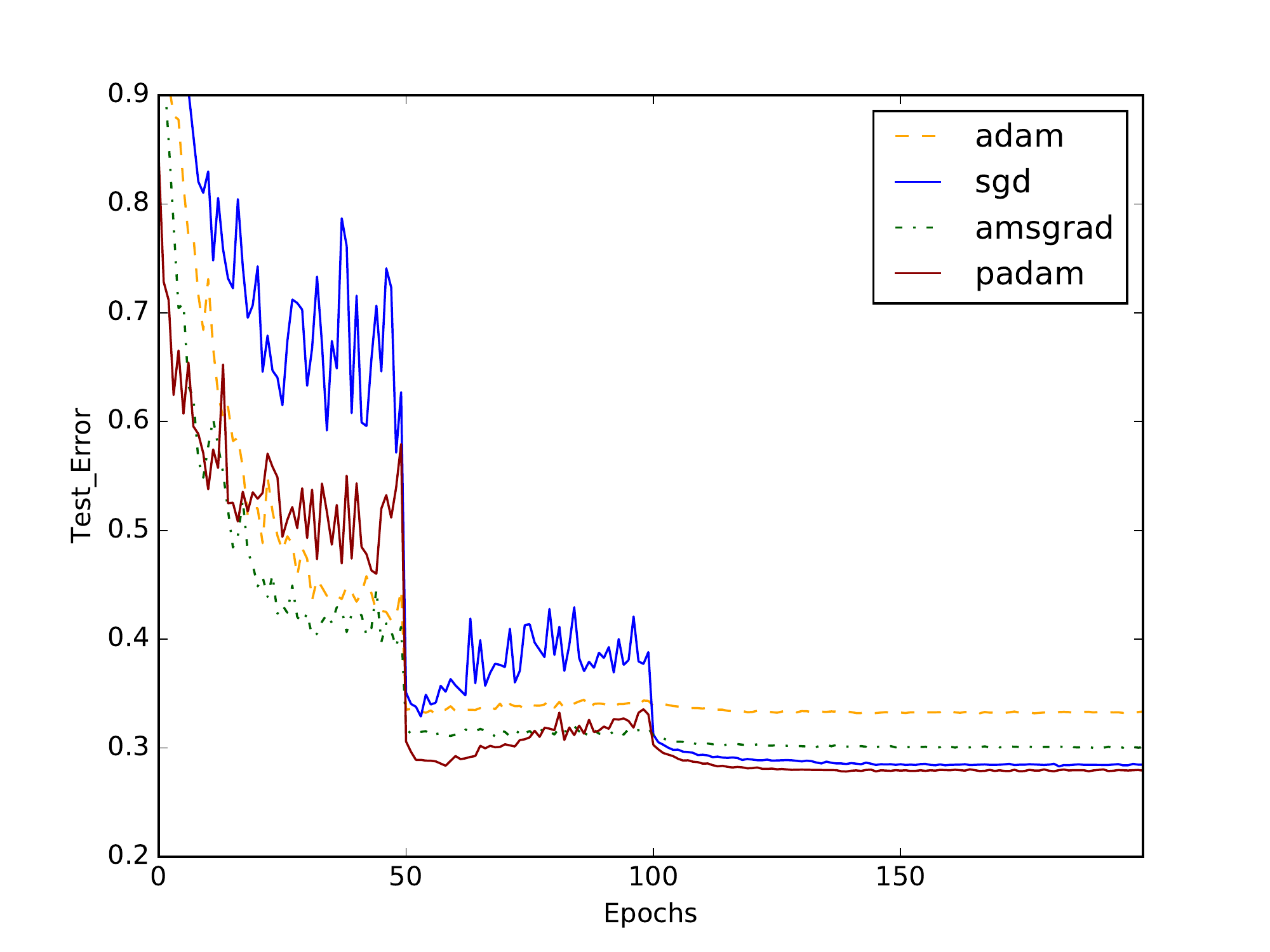}
\caption{Test Error for VGGNet}
\label{fig:subim22}
\end{subfigure}
\begin{subfigure}{0.5\textwidth}
\includegraphics[width=1\linewidth, height=4.4cm]{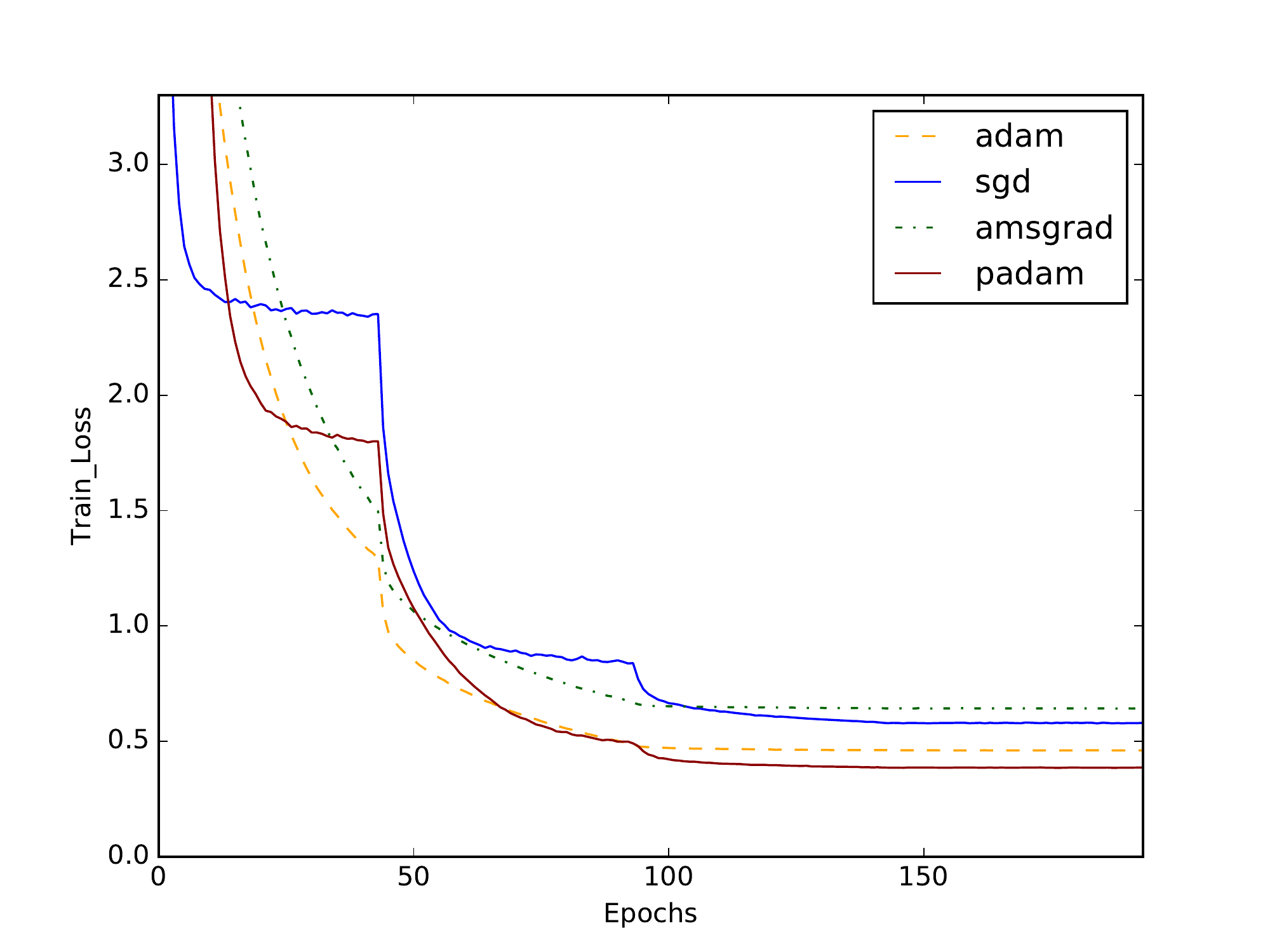}
\caption{Train Loss for Wide ResNet}
\label{fig:subim25}
\end{subfigure}
\begin{subfigure}{0.5\textwidth}
\includegraphics[width=1\linewidth, height=4.4cm]{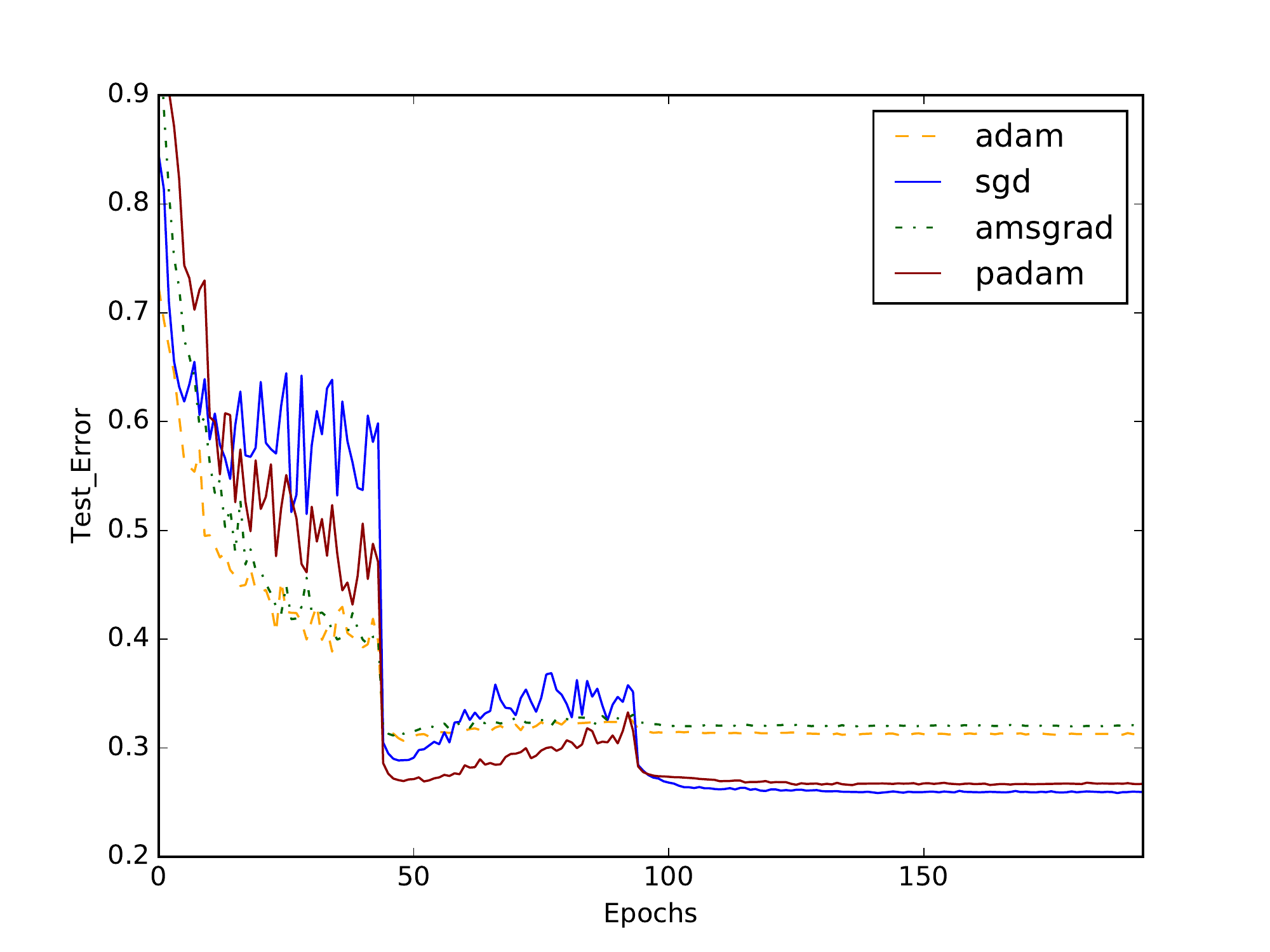}
\caption{Test Error for Wide ResNet}
\label{fig:subim26}
\end{subfigure}
\caption{Train loss and test error (top-1 error) of two CNN architectures on CIFAR-100.}
\label{fig:image24}
\end{figure}

\begin{table}[H]
\caption{Test Accuracy of VGGNet on CIFAR-100}
\label{sample-table2}
\begin{center}
\begin{tabular}{l|ccccc}
 \hline 
\multicolumn{1}{l}{\bf Methods}  &\multicolumn{1}{c}{\bf 50th epoch} &\multicolumn{1}{c}{\bf 100th epoch} &\multicolumn{1}{c}{\bf 150th epoch} &\multicolumn{1}{c}{\bf 200th epoch}
\\ \hline 
SGD Momentum &37.29 &61.18 &71.55 &71.54 \\
Adam &55.44 &65.67 &66.70 &66.65 \\
Amsgrad &{\bf 58.85} &{\bf 68.21}  &69.94 &69.95 \\
Padam &42.05 &66.92 &{\bf 72.04} &{\bf 72.08} \\ \hline
\end{tabular}
\end{center}
\end{table}

\begin{figure}[H]
\begin{subfigure}{0.5\textwidth}
\includegraphics[width=1\linewidth, height=5cm]{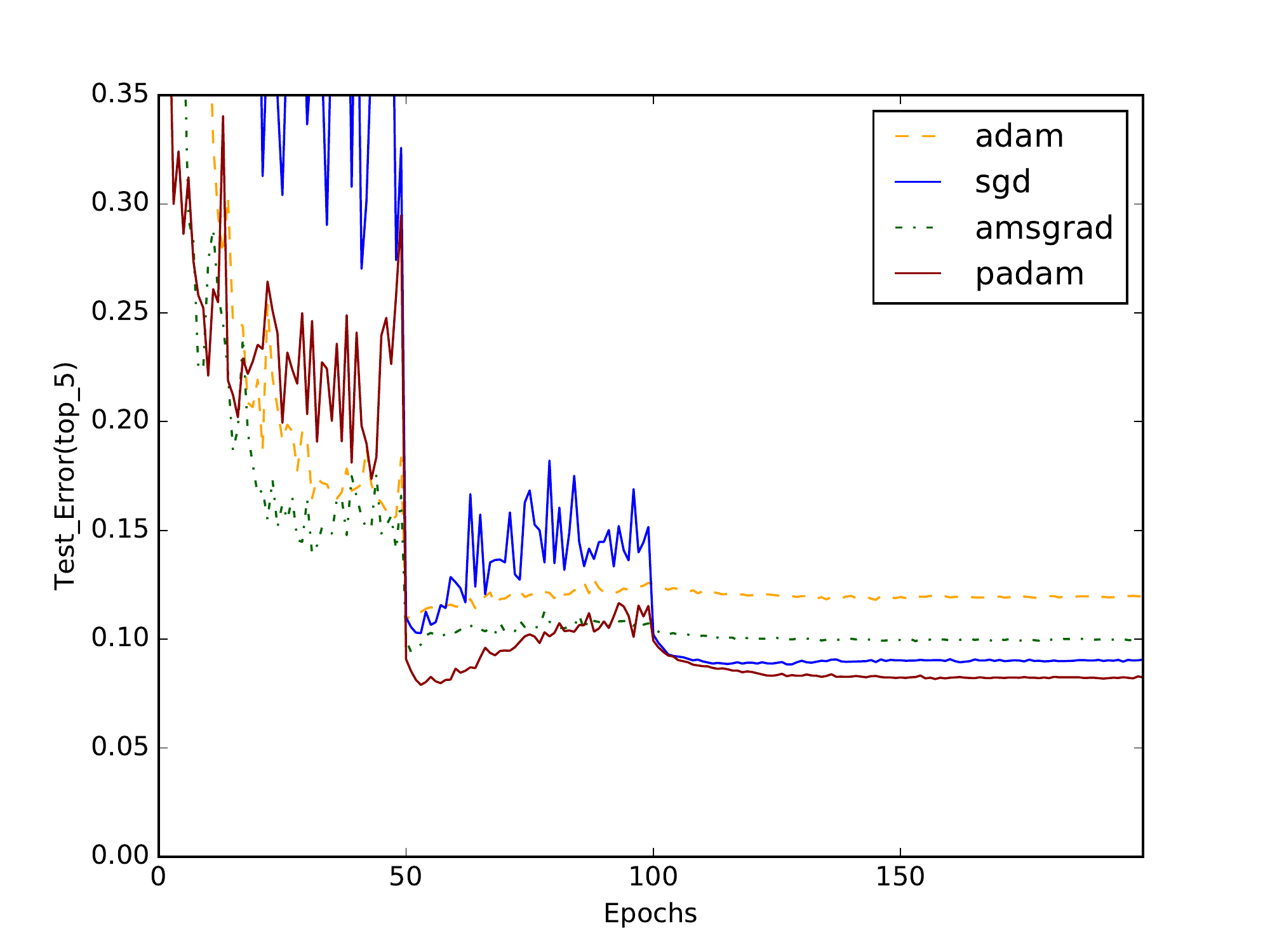}
\caption{Top-5 Error for VGGNet}
\label{fig:subim27}
\end{subfigure}
\begin{subfigure}{0.5\textwidth}
\includegraphics[width=1\linewidth, height=5cm]{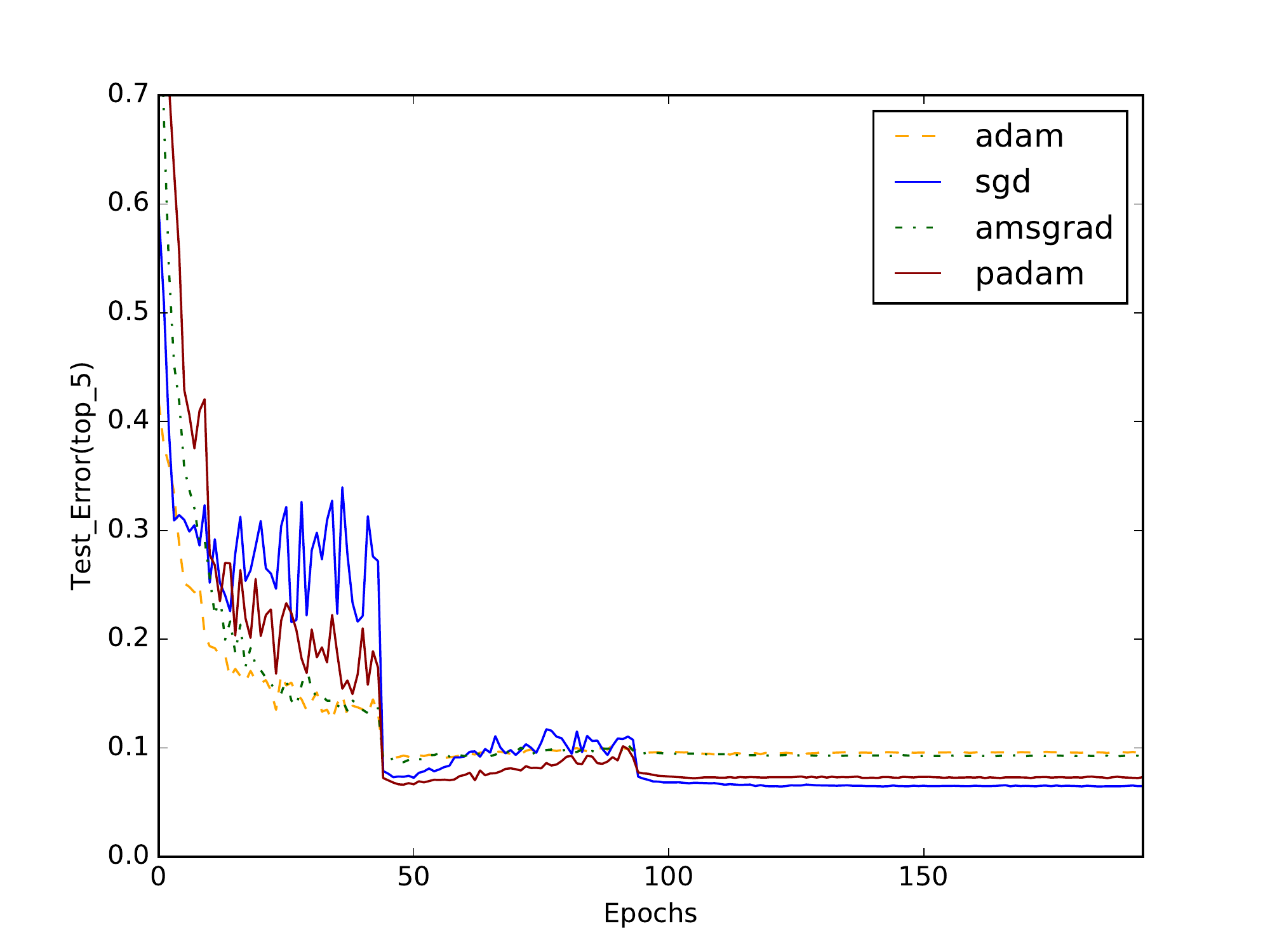}
\caption{Top-5 Error for Wide ResNet}
\label{fig:subim29}
\end{subfigure}
\caption{Top-5 error for two CNN architectures on CIFAR-100.}
\label{fig:image25}
\end{figure}

\begin{figure}[h]
 
\begin{subfigure}{0.5\textwidth}
\includegraphics[width=1\linewidth, height=5cm]{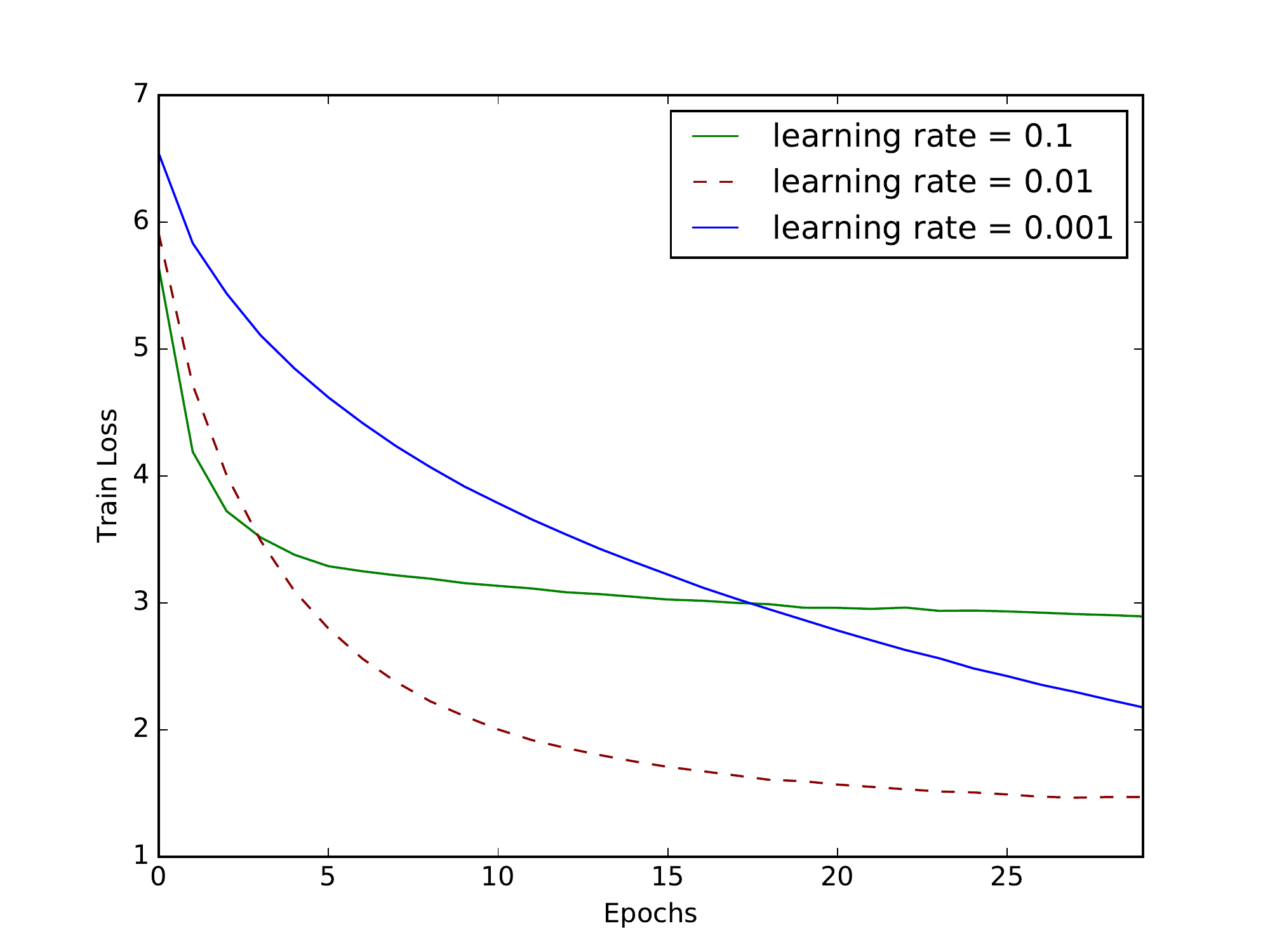}
\caption{Train Loss with p = 0.25}
\label{fig:subim30}
\end{subfigure}
\begin{subfigure}{0.5\textwidth}
\includegraphics[width=1\linewidth,height=5cm]{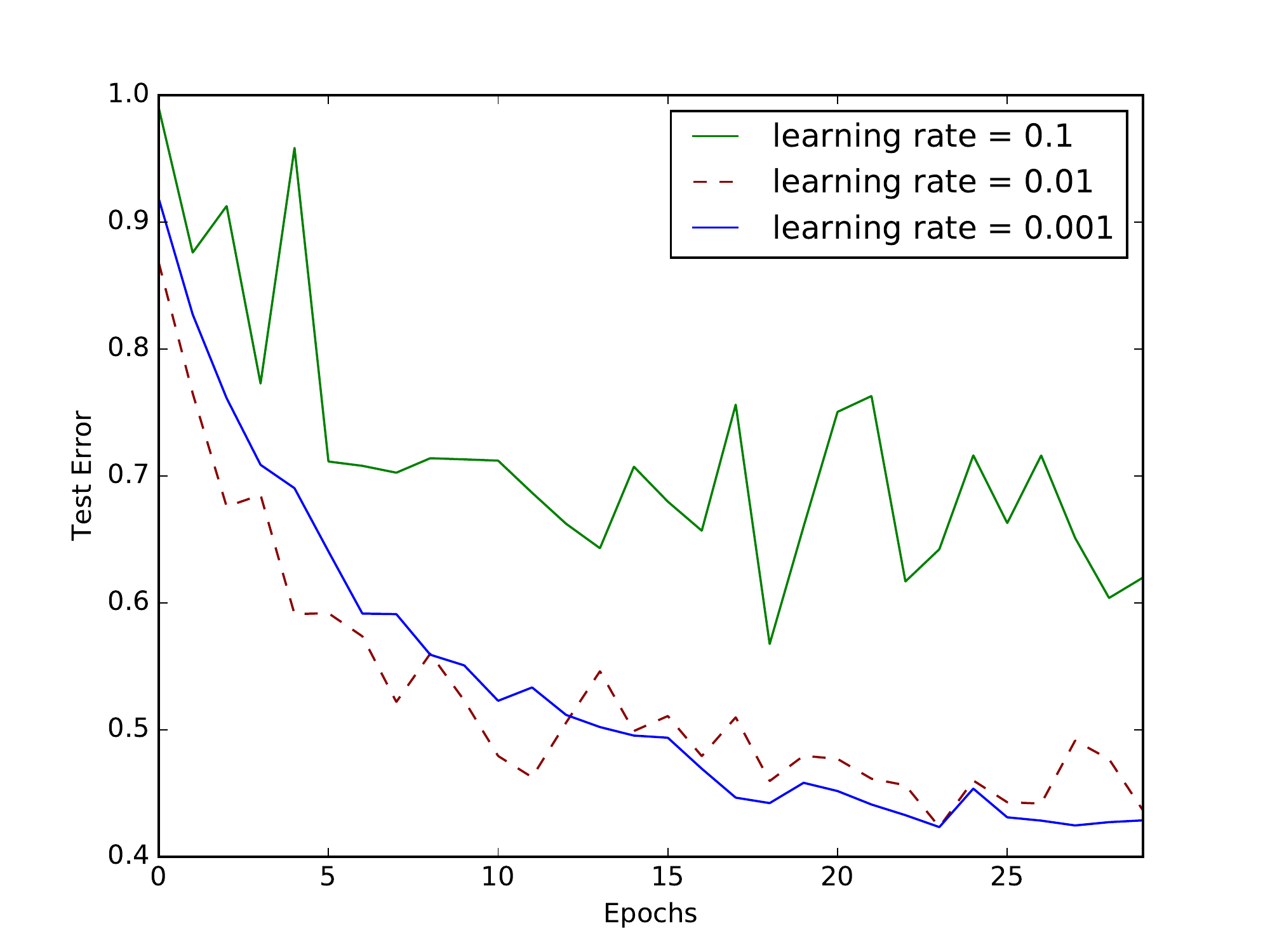}
\caption{Test Error with p = 0.25}
\label{fig:subim31}
\end{subfigure}
\begin{subfigure}{0.5\textwidth}
\includegraphics[width=1\linewidth, height=5cm]{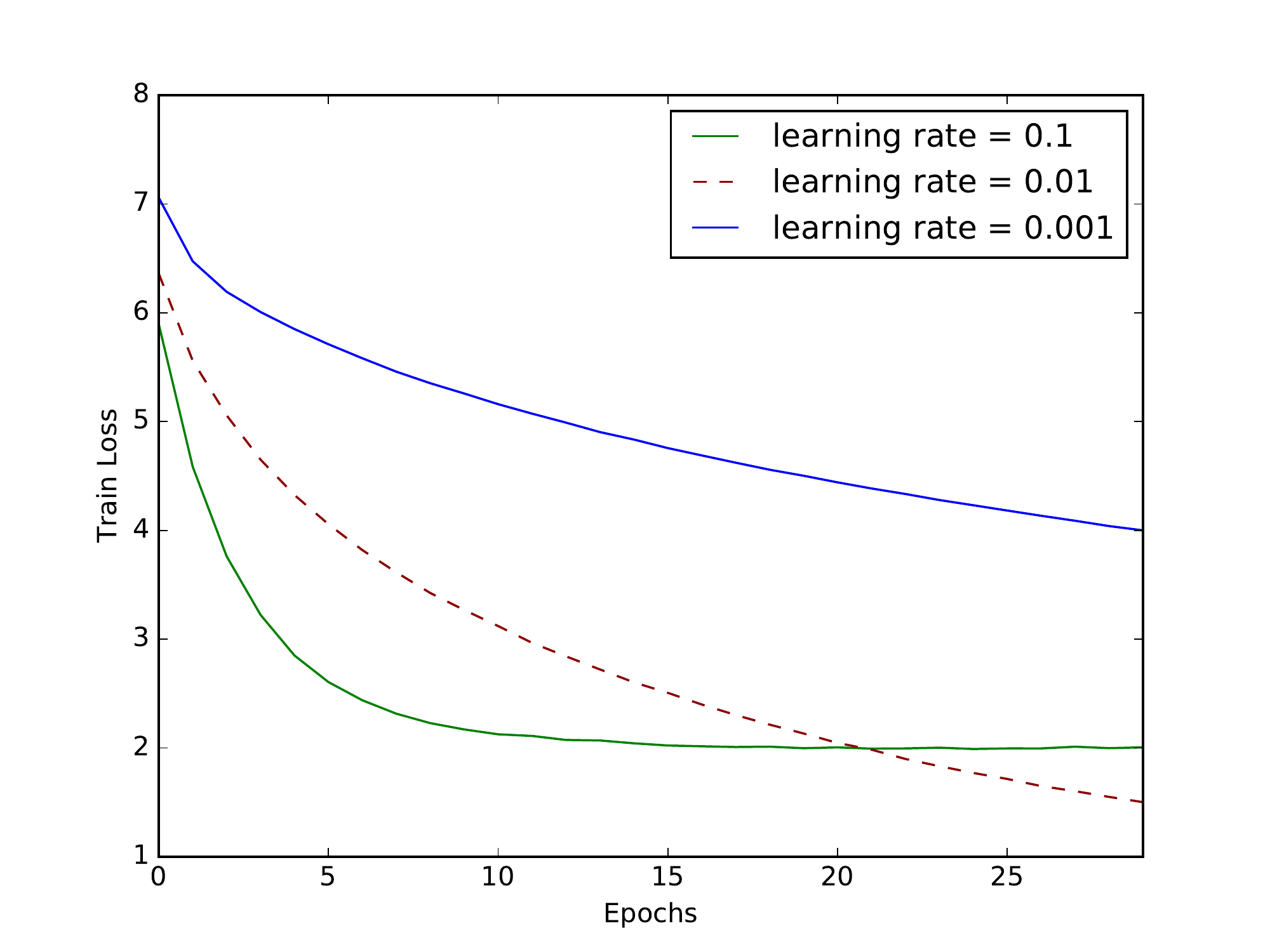}
\caption{Train Loss with p = 0.125}
\label{fig:subim32}
\end{subfigure}
\begin{subfigure}{0.5\textwidth}
\includegraphics[width=1\linewidth, height=5cm]{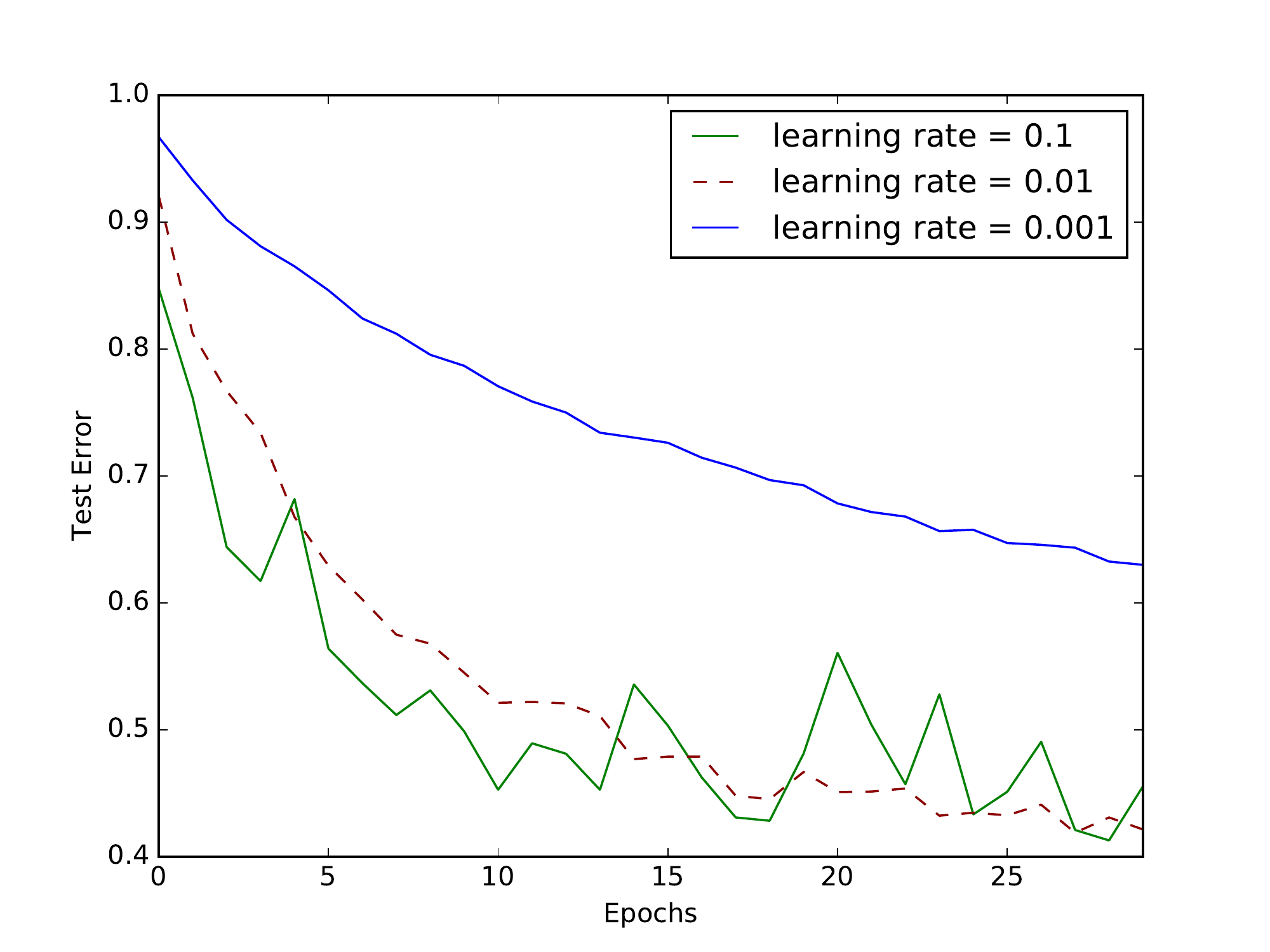}
\caption{Test Error with p = 0.125}
\label{fig:subim33}
\end{subfigure}
\begin{subfigure}{0.5\textwidth}
\includegraphics[width=1\linewidth, height=5cm]{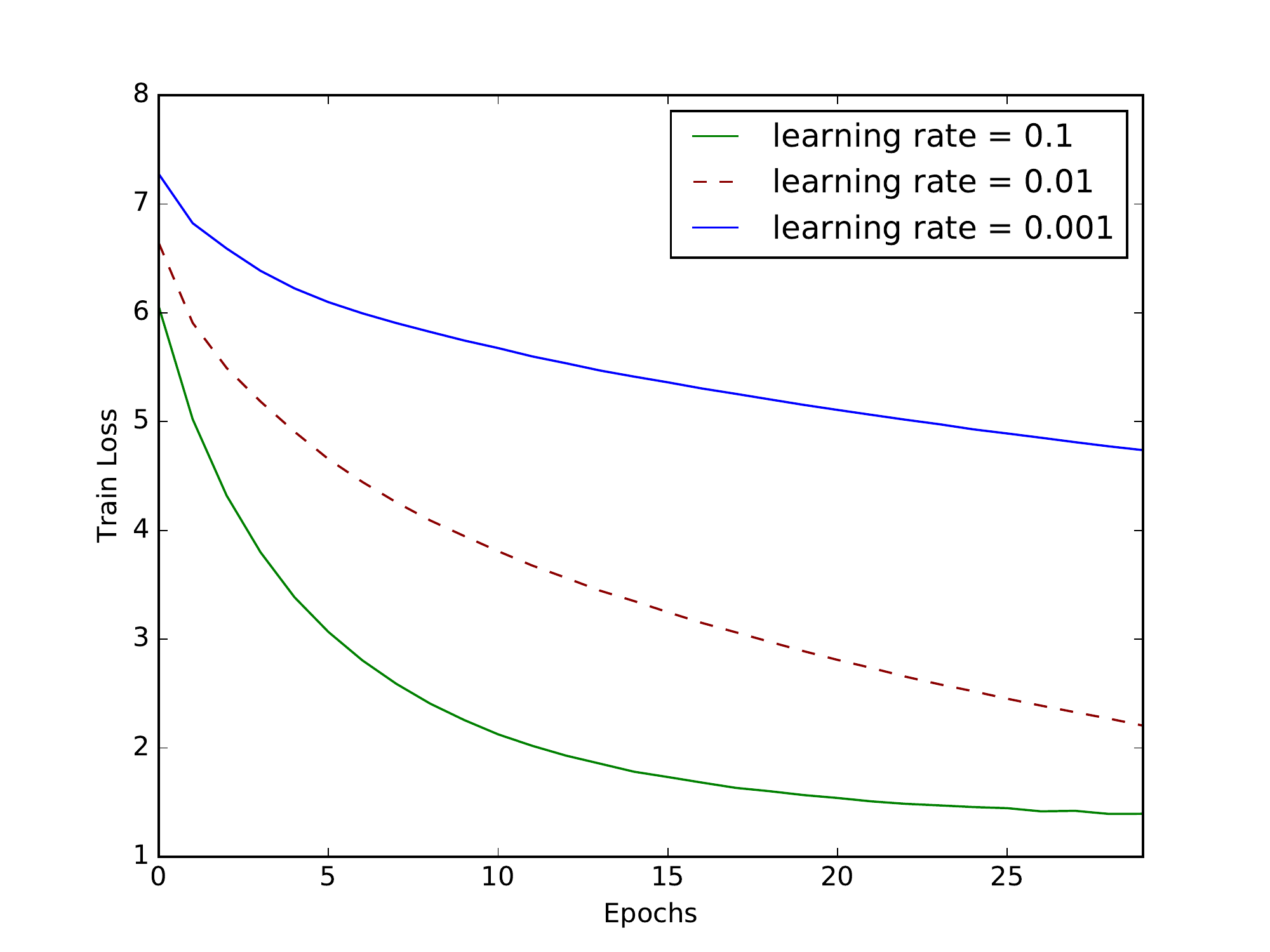}
\caption{Train Loss with p = 0.0625}
\label{fig:subim34}
\end{subfigure}
\begin{subfigure}{0.5\textwidth}
\includegraphics[width=1\linewidth, height=5cm]{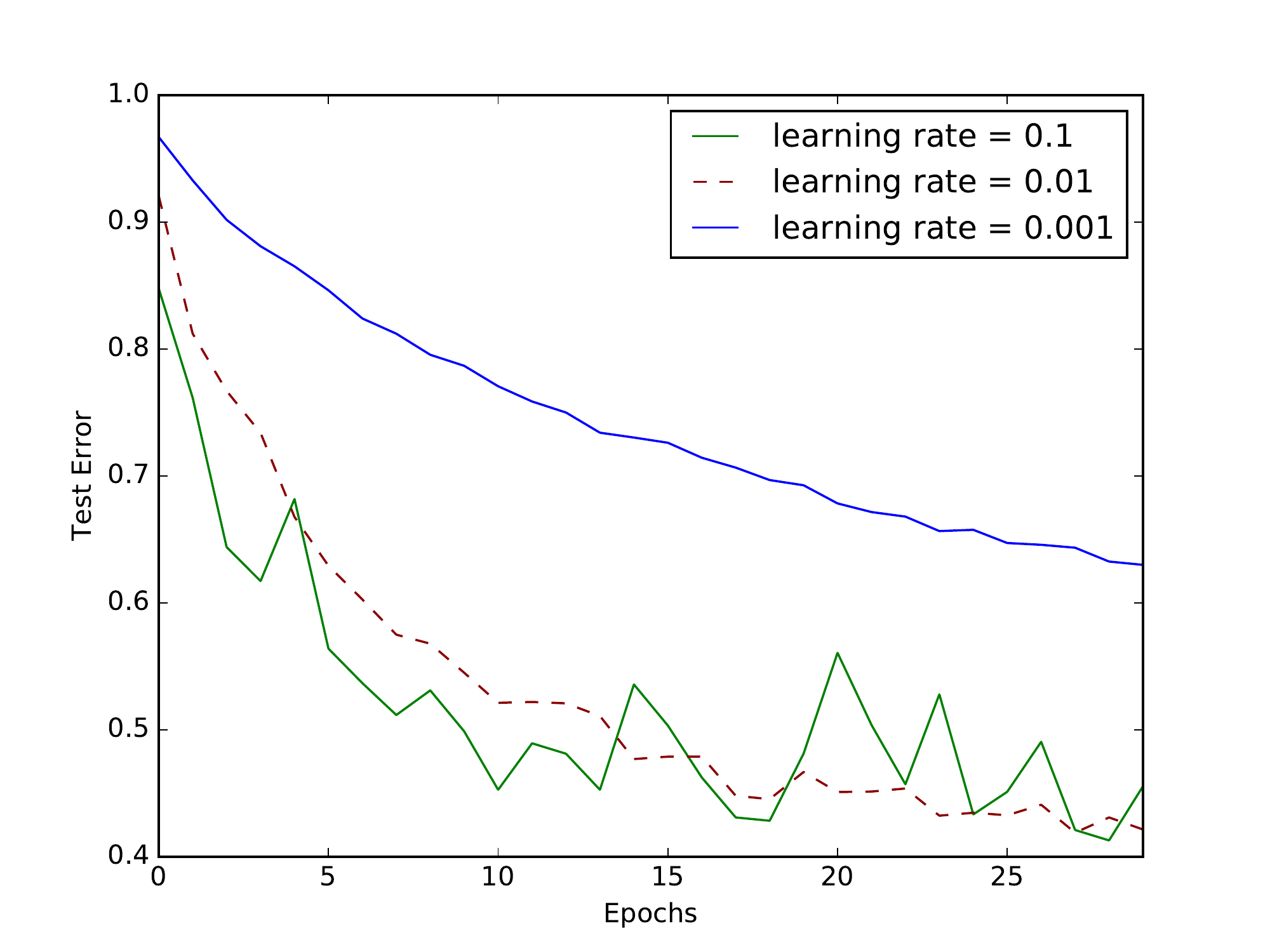}
\caption{Test Error with p = 0.0625}
\label{fig:subim35}
\end{subfigure}
\caption{Performance comparison of Padam with different choices of learning rate for three different fixed values of p (0.25, 0.125, 0.0625) for ResNet on CIFAR-100 dataset.}
\label{fig:image3}
\end{figure}

\end{document}